\documentclass{article}
\usepackage{sty/preamble}
\usepackage[utf8]{inputenc}

\title{TMI! Finetuned Models Leak Private \\ Information from their Pretraining Data}
\author{John Abascal\thanks{Khoury College of Computer Sciences, Northeastern University. Supported by the Khoury PhD Fellowship.  \texttt{abascal.j@northeastern.edu}} \and Stanley Wu\thanks{Khoury College of Computer Sciences, Northeastern University.  Supported by an REU supplement for NSF award CCF-1750640.  \texttt{stanleykywu@gmail.com}} \and Alina Oprea\thanks{\noindent Khoury College of Computer Sciences, Northeastern University.   Supported by NSF awards CNS-2120603 and CNS-2247484. \texttt{a.oprea@northeastern.edu}} \and Jonathan Ullman\thanks{Khoury College of Computer Sciences, Northeastern University.  Supported by NSF awards CNS-2120603, CNS-2232692, and CNS-2247484.  \texttt{jullman@ccs.neu.edu}.}}
\date{}
\usepackage[sortcites, maxbibnames=99]{biblatex}

\newcommand{\attackname}{\textbf{TMI}}

\newcommand{\revision}[1]{{#1}}

\begin{document}

\maketitle

% \large\textbf{Writing Tasks}

% \begin{itemize}
%     \item Intro
%     \item Related Work
%     \item Current attack algorithm
%     \item Current results
% \end{itemize}

% \large\textbf{Experimental Tasks}
% \begin{itemize}
%     \item Try global metaclassifier
%     \item Use adapted version of LiRA as a baseline (Done)
%     \item Sentiment classification
%     \item Disjoint vision tasks (Done)
%     \item “Zooming in” on a specific subset of CIFAR10 (Done)
%     \item DP fine tuning 
%     \item Fine tuning each layer (or different FT strategies)
% \end{itemize}

% \pagebreak 

\begin{abstract}
    Transfer learning has become an increasingly popular technique in machine learning as a way to leverage a pretrained model trained for one task to assist with building a finetuned model for a related task. This paradigm has been especially popular for \emph{privacy} in machine learning, where the pretrained model is considered public, and only the data for finetuning is considered sensitive.  However, there are reasons to believe that the data used for pretraining is still sensitive, making it essential to understand how much information the finetuned model leaks about the pretraining data. In this work we propose a new membership-inference threat model where the adversary only has access to the finetuned model and would like to infer the membership of the pretraining data. To realize this threat model, we implement a novel metaclassifier-based attack, \attackname, that leverages the influence of memorized pretraining samples on predictions in the downstream task. We evaluate \attackname\ on both vision and natural language tasks across multiple transfer learning settings, including finetuning with differential privacy. \footnote{Open-source implementation of \attackname\ : \url{https://github.com/johnmath/tmi-pets24}} Through our evaluation, we find that \attackname\ can successfully infer membership of pretraining examples using query access to the finetuned model. 
\end{abstract}

%  Transfer learning has become an increasingly popular technique in machine learning as a way to leverage a pretrained model trained for one task to assist with building a finetuned model for a related task. This paradigm has been especially popular for \emph{privacy} in machine learning, where the pretrained model is considered public, and only the data for finetuning is considered sensitive.  However, there are reasons to believe that the data used for pretraining is still sensitive, making it essential to understand how much information the finetuned model leaks about the pretraining data.  In this work we propose a novel metaclassifier-based membership-inference attack, \attackname, and show that it can successfully infer membership of pretraining examples, using only the finetuned model.

%\keywords{private machine learning, transfer learning, membership inference}

%     \includegraphics[width=0.4\textwidth]{figures/threat_model_gradient.pdf}
%     \caption{Our New Membership-Inference Threat Model}
%     \label{fig:ft_parameters}
\maketitle

% Motivating Ex 1: Explain the specific scenario where OpenAI/Google finetunes a model for another organization, but never gives them access to the whitebox model https://cloud.google.com/vertex-ai/docs/generative-ai/models/tune-models

% Motivating Ex 2: Explain the scenario where 

\section{Introduction}

Transfer learning has become an increasingly popular technique in machine learning as a way to leverage a model trained for one task to assist with building a model for a related task.  Typically, we begin with a large \emph{pretrained model} trained with abundant data and computation, and use it as a starting point for training a \emph{finetuned model} to solve a new task where data and computation is scarce.  This paradigm has been especially popular for \emph{privacy} in machine learning~\cite{papernot-2020-making, yu-2022-dpllm, li-2022-llmstrongdp, bu-2023-dpbiasft, he-2022-groupwisedp, ganesh-2023-pretraining}, because the data for pretraining is often considered public and thus the pretrained model provides a good starting point before we even have to touch sensitive data. 

Although the data used to pretrain large models is typically scraped from the Web and publicly accessible, there are several reasons to believe that this data is still sensitive~\cite{2022-tramer-position}. For example, personal data could have been published without consent by a third party who they trusted to keep their data private, and even ubiquitous and thoroughly examined pretraining datasets like ImageNet contain sensitive content~\cite{quach-2019-imagenet, yang-2021-imagenetfaces}. \revision{ Beyond the privacy risks associated with individuals in the pretraining set, companies who utilize or sell finetuned models may also be at risk for privacy leakage. Consider the following example:

\begin{example} Companies have large, web scraped datasets that are proprietary and remain internal (e.g., Google’s  JFT-300~\cite{JFT300}). These datasets are used to train models that can be finetuned by individual teams within the company for their specific needs. These pretrained models are also hosted as a service where smaller companies can receive a finetuned model without ever seeing the pretrained model itself. For example, Google's Vertex AI~\cite{2023-vertexai} allows smaller companies and individuals to upload their data and receive access to query the finetuned model as an endpoint. When these finetuned models are hosted publicly, they may leak sensitive information about the proprietary pretraining datasets on which they were trained. \end{example} 
% This example leads us to Q1.

% Additionally, companies have begun using proprietary internal datasets scraped from the Web that are up to $300\times$ the size of ImageNet \revision{(e.g. JFT-300 \cite{JFT300})}, making it imperative to understand the privacy risks posted by models pretrained on these ostensibly public datasets.

% Moreover, large organizations, \revision{such as OpenAI, Microsoft, and Google}, with the resources to pretrain large models as providing them as a service to smaller organizations without these resources to use as a starting point for transfer learning~\cite{openai-api, microsoft-api, google-api, 2023-vertexai}. Thus, even when organizations keep the pretraining model internal, sensitive information contained in the pretrained model may still leak out via these finetuned models.  

Thus, a central question we attempt to understand in this work is: {\em How much sensitive information does a finetuned model reveal about the data that was used for \emph{pretraining}?} We attempt to answer this question in both the settings where privacy preserving techniques have and have not been used to finetune the pretrained model. Examining these two settings leads to another research problem: Given that pretraining datasets have been shown to contain sensitive information~\cite{quach-2019-imagenet, yang-2021-imagenetfaces, carlini-2020-extraction_llm}, using the privacy preserving finetuning techniques described in prior work~\cite{papernot-2020-making, yu-2022-dpllm, li-2022-llmstrongdp, bu-2023-dpbiasft, he-2022-groupwisedp, ganesh-2023-pretraining} may not provide meaningful privacy guarantees in practice. }

% Additionally, we consider the privacy risks associated with pretrained models that are finetuned using privacy preserving techniques, such as \textit{differential privacy}~\cite{dwork-2006-dmns}. 

\revision{
% Setup for motivating example involving private finetuning. 
\begin{example} 
Using the thought experiment from \cite{2022-tramer-position}, consider a large, pretrained model, owned by Company A, that contains an individual's sensitive data record. Suppose that this pretrained model is finetuned by Company B using \textit{differential privacy}~\cite{dwork-2006-dmns} with $(\varepsilon=0.5, \: \delta=10^{-5})$ on a sensitive downstream task. Considering that the open-source variants of these models' pretraining datasets can exceed 5 TB (over 1 trillion tokens) in size~\cite{2023-llama-dataset}, it is likely that any given individual's data record can be present in both the pretraining and finetuning datasets. Thus, because differential privacy is not necessarily robust to preprocessing, the privacy guarantee from finetuning may not hold for individuals whose data record is present in both datasets. 
\end{example} 

To this end, we will also attempt to answer the following question: \textit{Does using differential privacy during finetuning always provide its stated privacy guarantee?}
% This example leads us to Q2.
}

We study \revision{these questions} via \textit{membership-inference (MI) attacks}.  A MI attack allows an adversary with access to the model to determine whether or not a given data point was included in the training data.  These privacy attacks were first introduced by Homer et al.~\cite{homer-2008-mi} in the context of genomic data, formalized and analyzed statistically by Sankararaman et al.~\cite{SankararamanOJH09} and Dwork et al.~\cite{DworkSSUV15}, and later applied in machine learning applications by Shokri et al.~\cite{shokri-2016-mi}.  MI attacks have been extensively studied in machine learning applications, such as computer vision \cite{carlini-2022-lira}, contrastive learning~\cite{2021-liu-encodermi}, generative adversarial networks~\cite{chen-2019-ganmi}, and federated learning~\cite{nasr-2019-federatedmia}.  The success of MI attacks makes it clear that the pretrained model will leak information about the pretraining data.  However, the process of finetuning the model will obscure information about the original model, and there are no works that study MI attacks that use the \emph{finetuned model} to recover \emph{pretraining data}.

We create a novel, metaclassifier-based membership-inference attack, \textbf{T}ransfer \textbf{M}embership \textbf{I}nference (\attackname) to circumvent the challenges that arise when trying to adapt prior attacks to asses privacy leakage in this new setting where the adversary has query access only to the finetuned model. The goal of our new membership-inference adversary is to infer whether or not specific individuals were included in the pretraining set of the finetuned machine learning model. This setting stands in contrast to prior membership-inference attacks, as it restricts the adversary from directly querying the model trained on the specific dataset they wish to perform membership-inference on. State-of-the-art, black-box MI attacks rely on a model's prediction confidence with respect to the ground truth label, but the finetuned model does not necessarily have the ground truth label in its range. Thus, our attack leverages how individual samples from pretraining influence predictions on the downstream task by observing entire prediction vectors from the finetuned model. More concretely, \attackname\ constructs a dataset of prediction vectors from queries to finetuned shadow models in order to train a metaclassifier that can infer membership.

%Whereas prior membership inference attacks attempt to utilize the influence that a single individual's data entry has on the prediction of the correct label, our attack is designed to be aware of the generic features learned in the pretraining process by incorporating information from the target model's predictions on all possible labels. By making this key observation, it becomes evident that privacy leakage exists in the transfer learning setting, whether or not privacy preserving techniques are used in the process of transferring a model to a downstream task. \ju{I can't make any sense of this high level description.}

\revision{We comprehensively evaluate \attackname\ on pretrained CIFAR-100~\cite{CIFAR} and Tiny ImageNet~\cite{tiny-imagenet} vision models, transferred to multiple downstream tasks. In our experiments with Tiny ImageNet, we evaluate the ability of \attackname\ to infer membership on models finetuned on Caltech 101~\cite{caltech-101}. Our pretrained CIFAR-100 models are finetuned on three downstream datasets of varying similarity to the pretraining data. In order of similarity to CIFAR-100, we evaluate \attackname\ on models finetuned on a coarse-labeled version of CIFAR-100, CIFAR-10~\cite{CIFAR}, and the Oxford-IIT Pet dataset~\cite{pets}. We also evaluate an extension of \attackname\ on finetuned versions of publicly available large language language models, which are pretrained on WikiText-103~\cite{wikitext103}}. To measure the success of our attack we use several metrics, such as AUC and true positive rates at low false positive rates. To demonstrate the prevalence of privacy leakage with respect to pretraining data in finetuned models, we run \attackname\ on target models with different finetuning strategies and settings with limited adversarial capabilities. We compare our results to both a simple adaptation of the likelihood ratio attack \cite{carlini-2022-lira} to the transfer learning setting and a membership inference attack that has direct access to the pretrained model.

\myparagraph{Our Contributions} We summarize our main contributions to the study of membership-inference attacks as follows:
\begin{itemize}
    \item We investigate privacy leakage in the transfer learning setting, where machine learning models are finetuned on downstream tasks with and without differential privacy.
    \item We introduce a new threat model, where the adversary only has query access to the finetuned target model.
    \item We propose a novel membership-inference attack, \attackname, that leverages all of the information available to the black-box adversary to infer the membership status of individuals in the pretraining set of a finetuned machine learning model.
    \item We provide theoretical results for membership-inference attacks on mean estimation to support and explain our findings. 
    \item We evaluate our attack across \revision{four} vision datasets of varying similarity to the \revision{two pretraining tasks} and several different transfer learning strategies. We show that there is privacy leakage even in cases where the pretraining task provides little benefit to the downstream task or the target model was finetuned with differential privacy. \revision{We also show that membership in the pretraining dataset can lead to unexpected privacy leakage when finetuning with differential privacy.}
    \item We study privacy leakage of finetuned models in the natural language domain by evaluating our attack on \revision{two finetuned versions} of a publicly available foundation model.
\end{itemize}

% Might not need to include TransMIA in the intro. We can talk about it in the related work, but it might not be so relevant because a white box attack on the pretrained part of an ML model is the same as a white box MIA in the standard setting

\section{Background and Related Work}

We provide the necessary background on machine learning, privacy in machine learning, and related work on existing inference attacks.

\subsection{Machine Learning Background and Notation}

% Explain how classification models learn class probabilities
In our attacks, we assume that the target models are classifier neural networks trained in a supervised manner.  A neural network classifier with parameters $\theta$ is a function, $f_{\theta} : \mathcal{X} \to [0,1]^{K}$ that maps data points $x \in \mathcal{X}$ to a probability distribution over $K$ classes. In the supervised learning setting, we are given a dataset of labeled $(x, y)$ pairs $D$ drawn from an underlying distribution $\mathbb{D}$ and a training algorithm $\mathcal{T}$. The parameters of the neural network are then learned by running the training algorithm over the dataset, which we will denote $f_{\theta} \gets \mathcal{T}(D)$. A popular choice for the training algorithm is stochastic gradient descent (SGD), which minimizes a loss function $\mathcal{L}$ over the labeled dataset $D$ by iteratively updating the models parameters $\theta$:

$$
    \theta_{i+1} \gets \theta_{i} - \frac{\eta}{m} \sum_{(x,y) \in D}{\nabla_{\theta} \mathcal{L}(f_{\theta}(x), y)}
$$

\noindent where $m$ is the dataset size, $\eta$ is a tunable parameter called the learning rate. In our setting, we define the loss function $\mathcal{L}$ to be the cross-entropy loss:

$$
    \mathcal{L}(f_{\theta(x)}, y) = - \sum^{K}_{j=1} \mathbbm{1}_{\{ j = y\}} \log(p_{j})
$$

\noindent where $p_{j}$ is the model's prediction probability for class $j$.

\subsubsection{Scaling Model Confidences} \hfill \label{subsection:logit_scaling}

% Include how confidences are rescaled to logits
The classifier models we consider output a vector of probabilities, $\vec{y}$, where each entry $y_{i}$ corresponds to the \textit{model's prediction confidence} with respect to label, $i$. This is done by applying the $\softmax$ activation function to the model's final layer. Given a vector of logits, $\vec{z}$ (i.e. the model's final layer), we define $\softmax(\vec{z}) : \R^K \to (0,1)^K$
$$
    y_i = \softmax(\vec{z})_i = \frac{e^{z_i}}{\sum_{j=1}^{K} e^{z_j}}
$$
where $K$ is the number of possible classes. 

% In our setting, the adversary does not have direct access to the model's logits, but they would like to recover the model's logits from the predicted probability vector. Solving for $z_i$ when we are given $\sigma(\vec{z})_i$

% \begin{align*}
%     \log(y_{i}) &= \log\Big(\frac{e^{z_i}}{\sum_{j=1}^{K} e^{z_j}}\Big) \\
%     &= \log(e^{z_i}) - \log(\sum_{j=1}^{K} e^{z_j}) \\
%     &= z_i -  \log(\sum_{j=1}^{K} e^{z_j}) \\
%     z_i &= \log(y_{i}) + \log(\sum_{j=1}^{K} e^{z_j}) \\
% \end{align*}

% \noindent Because $\softmax$ is not invertible, the adversary necessarily needs access to the model's logits to learn $z_i$. 

Prior work \cite{carlini-2022-lira} has used the logit function, $\texttt{logit}(p) = \smash{\log(\tfrac{p}{1-p})}$, to scale model confidences. This scaling yields an approximately normally distributed statistic that can be used to perform a variety of privacy attacks \cite{carlini-2022-lira,chaudhari-2022-snap,tramer-2022-truthserum}. The logit function is obtained by inverting the \texttt{sigmoid} activation function, $\sigma(x) = \frac{1}{1+ e^{x}}$, which is a specific case of $\softmax$ being used for binary classification. 

Following the lead of prior work, we use $\phi$ to perform our model confidence scaling. We define model confidence scaling $\phi(\vec{y}): \R^K \to \R^K$ for a prediction vector, $\vec{y}$, as
\[
\phi(\vec{y}) = (\texttt{logit}(y_{1}),\dots,\texttt{logit}(y_{K}))
\]

\subsubsection{Transfer Learning} \hfill

% Explain feature extraction and how it is a common tecnique in cheaply finetuning deep learning models 

Feature extraction and updating a model's pretrained weights are popular transfer learning techniques used to improve a pretrained deep learning model's performance on a specific task. In the classification setting, feature extraction involves freezing a model's pretrained weights and using them to extract relevant features from input data, which are then fed into a linear layer for classification. This technique is useful when working with limited data or when the pretrained model has learned generalizable features that are useful for the target task. On the other hand, finetuning a model by updating its pretrained weights involves taking a pretrained model and training it on a new dataset, often with a smaller learning rate, to adapt it to the new task. This kind of finetuning is more suited for situations where the new task has similar characteristics, but not a direct correspondence, to the original pretraining task.

\subsubsection{Differential Privacy} \hfill

\textit{Differential Privacy} \cite{dwork-2006-dmns} is a mathematical definition of privacy that bounds the influence that any single individual in the training data has on the output of the model.  Specifically, an algorithm satisfies differential privacy if for any two datasets that differ on one individual's training data, the probability of seeing any set of potential models is roughly the same regardless of which dataset was used in training.

\begin{defn}
A randomized algorithm $\mathcal{M}$ mapping datasets to models satisfies \emph{$(\eps,\delta)$-differential privacy} if for every pair of datasets $X$ and $X'$ differing on at most one training example and every set of outputs $E$,
\begin{equation*}
    \Pr[\mathcal{M}(X) \in E] \leq e^{\varepsilon} \Pr[\mathcal{M}(X') \in E] + \delta
\end{equation*}
\end{defn}

\subsection{Related Work} 
% \subsubsection{Memorization in Deep Learning Models} \hfill \\

% As a result of this memorization, deep learning models tend to have higher prediction confidence on training data. This disproportionate confidence has been exploited to mount membership inference attacks \cite{carlini-2022-lira, shokri-2016-mi} and makes deep learning models highly susceptible to privacy attacks.

\subsubsection{Privacy Attacks on Machine Learning Models} \hfill 

Deep learning models have been shown to memorize entire individual data points, even in settings where the data points have randomly assigned labels \cite{zhang-2017-rethinking}. Prior work has demonstrated the ability of a wide class of deep learning models to perfectly fit training data while also achieving low generalization error \cite{belkin-2019-reconciling}. In fact, recent work \cite{feldman-2020-longtail, brown-2021-irrelevant, feldman-2020-memorize} has shown that memorization of training data may actually be necessary to achieve optimal generalization for deep learning models. As a result of this memorization, deep learning models tend to have higher prediction confidence on training data, which makes them highly susceptible to privacy attacks.

The most glaring violations of privacy in machine learning are reconstruction and training data extraction attacks. Early work in data privacy \cite{dinur-2003-reconstruction} showed that it is possible to reconstruct individuals' data in statistical databases with access to noisy queries. More recently, training data extraction attacks have been shown to be successful when mounted on a variety of deep learning models, including large language models~\cite{carlini-2020-extraction_llm} and computer vision models~\cite{carlini-2023-extraction_diffusion}.

Other attacks on machine learning models, such as membership-inference~\cite{shokri-2016-mi}, property inference~\cite{ateniese-2015-property_inference}, and attribute inference~\cite{2018-yeom-mioverfitting} attacks are more subtle privacy violations. These attacks exploit vulnerabilities in machine learning models to learn whether or not an individual was in the training set, global properties of the training dataset, and an individual's sensitive attributes. respectively. Recent versions of these attacks typically use a test statistic, such as loss \cite{ye-2022-enhanced_mi} and model prediction confidences~\cite{2018-yeom-mioverfitting, carlini-2022-lira, chaudhari-2022-snap}, to extract private information.

\subsubsection{Membership-Inference Attacks} \hfill

Membership-inference attacks~\cite{homer-2008-mi} aim to determine whether or not a given individual's data record was present in a machine learning model's training dataset. These attacks represent a fundamental privacy violation that has a direct connection to differential privacy. Mounting these attacks and learning whether or not an individual was part of a sensitive dataset can serve as the basis for more powerful attacks. For example, prior work has used MI as a step in extracting training data~\cite{carlini-2020-extraction_llm}. Because of their simplicity, MI attacks are also a popular way to audit machine learning models for privacy leakage~\cite{song-2018-auditingmi, tensorflow-2020-auditing, ye-2022-enhanced_mi}.

These attacks been extensively studied with two types of adversarial access: black-box query access and white-box access to the machine learning model's parameters~\cite{2020-klas-whitebox}. The query access setting has been more thoroughly studied, with attacks spanning several different machine learning domains, such as classification~\cite{carlini-2022-lira, 2018-yeom-mioverfitting, shokri-2016-mi, ye-2022-enhanced_mi, tramer-2022-truthserum}, natural language generation~\cite{carlini-2022-lira, tramer-2022-truthserum}, and federated learning~\cite{nasr-2019-federatedmia}. Despite there being extensive work on black-box attacks and prior work on MI attacks on pretrained encoders~\cite{2021-liu-encodermi}, continuously updated models~\cite{jagielski-2023-updates}, and distilled models~\cite{jagielski-2023-students}, there are few works that explore MI in the transfer learning setting where a pretrained model is finetuned on a new task. Zou et al.~\cite{zhang-2020-transfer} study MI attacks that target individuals in the finetuning dataset, and Hidano et al.~\cite{hidano-2020-transmia} explore ways in which an adversary can leverage control over the transfer learning process to amplify the success of MI attacks on the original model. No works have studied black-box MI attacks on the pretraining dataset of a finetuned machine learning model.

\section{Threat Model} \label{sec:threatmodel}

\begin{figure}[]
    \centering
    \includegraphics[width=0.8\textwidth]{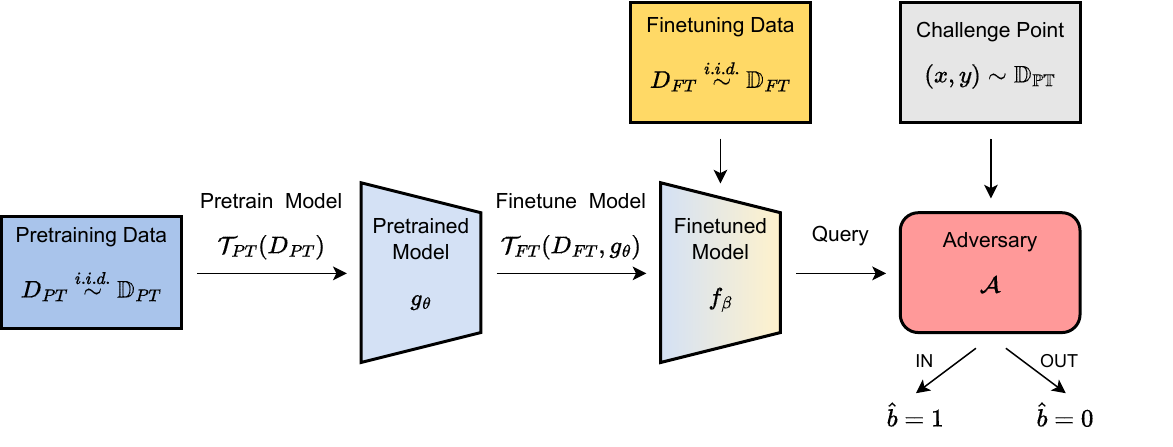}
    \caption{Our New Membership-Inference Threat Model.}
    \label{fig:threat_model}
\end{figure}

Our problem is to determine how much information a finetuned model reveals about the data used in the pretraining phase, and whether or not the finetuned model reveals strictly less information than the pretrained model.  For this work we study this question using the language of MI attacks~\cite{homer-2008-mi, SankararamanOJH09, DworkSSUV15}.   In the standard MI experiment and in our newly defined experiment, there is a machine learning model trained on some dataset, and a challenge point that is drawn from the same distribution as the training data.  The challenge point is either an element of the training data or an independent point drawn from the same distribution.  The attacker, who has access to the model and the challenge point, and knowledge of the distribution, tries to infer which of these two cases holds.  In our experiment we separate the construction of the machine learning model into a pretraining phase and a finetuning phase, where the finetuning phase is performed with different data, drawn from a possibly different distribution.  This finetuning phase introduces another layer of indirection that prevents the attacker from querying the original pretrained model, and thus potentially makes MI more difficult.  Formally, our threat model, visualized in Figure~\ref{fig:threat_model}, is described by the following game between a \textit{challenger} $\mathcal{C}$ and an \textit{adversary} $\mathcal{A}$:

%The threat model for membership inference on machine learning models first introduced by Shokri et al.~\cite{shokri-2016-mi} assumes that the adversary has direct query access to a target model. In this paper, we consider a variation of this setting where the adversary uses query access to a finetuned model to infer the membership status of points in the pretraining data. This means that the target model our adversary interacts with with has some intermediate layers that are not necessarily related to the pretraining set on which we would like to perform membership inference. To describe this new membership inference setting, we define the membership inference security game between a \textit{challenger} $\mathcal{C}$ and an \textit{adversary} $\mathcal{A}$  by slightly modifying the definition outlined by Carlini et al.~\cite{carlini-2022-lira}. \\

\noindent \textbf{MI Security Game with a Finetuned Target Model}
\begin{enumerate}
    \item The challenger receives a dataset $D_{PT}$ comprised of points sampled i.i.d. from some distribution $\mathbb{D}_{PT}$, and a pretrained model $g_{\theta} \gets \mathcal{T}_{PT}(D_{PT})$.
    
    \item The challenger draws i.i.d. samples from  another distribution $\mathbb{D}_{FT}$ to create a dataset $D_{FT}$ and finetunes the model on $D_{FT}$ using its pretrained weights, $\theta$, to obtain a new model $f_{\beta}  \gets \mathcal{T}_{FT}(D_{FT}, g_{\theta})$.
    
    \item The challenger randomly selects $b \in \{0, 1\}$. If  $b = 0$, the challenger samples a point $(x,y)$ from $\mathbb{D}_{PT}$ uniformly at random, such that $(x,y) \notin D_{PT}$. Otherwise, the challenger samples $(x,y)$ from $D_{PT}$ uniformly at random. 
    \item The challenger sends the point, $(x,y)$ to the adversary.
    
    \item The adversary, using the challenge point, sampling access to $\mathbb{D}_{PT}$ and $\mathbb{D}_{FT}$, and query access to $f_{\beta}$, produces a bit $\hat b$.
    
    \item The adversary wins if $b = \hat b$ and loses otherwise.
\end{enumerate} 

In our security game, we assume that the adversary has query access to the finetuned target model $f_{\beta}$ and knowledge of the pretraining data distribution $\mathbb{D}_{PT}$. Because we will be training \textit{shadow models}~\cite{shokri-2016-mi} to perform our MI attack, the adversary also requires knowledge of the underlying distribution from which the finetuning dataset is sampled, $\mathbb{D}_{FT}$, and knowledge of the target model's architecture and training algorithm.  Although MI attacks vary in what they assume about the distribution and training algorithm~\cite{DworkSSUV15}, some degree of knowledge is necessary, so we consider full knowledge as a reasonable starting point.  The knowledge we assume is the same as many other works on MI (e.g.~\cite{shokri-2016-mi, carlini-2022-lira, tramer-2022-truthserum, 2018-yeom-mioverfitting, sablayrolles-2019-whitebox, jagielski-2023-students, 2021-liu-encodermi}). 
%As in most other works on MI, 
We also assume that the adversary's queries to the target model return numerical confidence scores for each label rather than just a single label, similar to prior privacy attacks~\cite{shokri-2016-mi,2018-yeom-mioverfitting,tramer-2022-truthserum,carlini-2022-lira}.

% Should we include the assumption of soft labels over hard labels?
% Can we make a hard label version of the attack?

\section{Methodology}

In this section, we will propose attacks that follow the threat model defined in Section~\ref{sec:threatmodel}. \revision{First, we will motivate our attack with theoretical results for membership-inference attacks under distribution shift. Then, we will provide a simple adaptation of an existing MI attack and describe issues that arise when trying to incorporate more information about target model queries into an attack implementation.} Lastly, we will detail our metaclassifier-based approach to performing black-box MI attacks on finetuned models.
% \begin{algorithm}
% \caption{\texttt{train\_shadow\_model}$(x,\: b)$ \\ Our shadow model training procedure considers both the pretraining and finetuning phases to mimic the behavior of the target model on a challenge point.}\label{alg:sm_training}
%     \begin{algorithmic}[1]
%         \Require A challenge point $x \leftarrow \mathbb{D}_{PT}$, and query access to both $\mathbb{D}_{PT}$ and $\mathbb{D}_{FT}$, and a boolean value, $b$, to denote whether the shadow model contains the challenge point in its training set.
%         \State Sample $\tilde{D}_{PT}$ i.i.d. using query access to $\mathbb{D}_{PT}$ such that $\tilde{D}_{PT}$ does not contain $x$
%         %  Pretrain a shadow model \textit{without} the challenge point
%         \If{$b = 0$}
%             \State $ g \leftarrow \mathcal{T}(\tilde{D}_{PT})$

%         %  Pretrain a shadow model \textit{with} the challenge point
%         \ElsIf{$b=1$}
%             \State $g \leftarrow \mathcal{T}(\tilde{D}_{PT} \cup \{x\})$
%         \EndIf
%         \State Sample $\tilde{D}_{FT}$ i.i.d. using query access to $\mathbb{D}_{FT}$ 
%         \State $f \leftarrow \mathcal{T}(g, \: \tilde{D}_{FT})$ \Comment{Finetune $g$ on $\tilde{D}_{FT}$}
%         \Return $f$
%     \end{algorithmic}
% \end{algorithm}

\begin{algorithm}
\caption{\texttt{train\_shadow\_models}$(x,\: b)$ \\ Our shadow model training procedure considers both the pretraining and finetuning phases to mimic the behavior of the target model on a challenge point.}\label{alg:sm_training}
    \begin{algorithmic}[1]
        \Require Query access to both $\mathbb{D}_{PT}$ and $\mathbb{D}_{FT}$ and a fixed dataset size $S = \frac{1}{2}\vert \mathbb{D}_{PT} \vert$
        
        \State $\text{models} \gets \{\}$
        \State $\text{datasets} \gets \{\}$
        \For{$N$ times}
            \State Draw $S$ i.i.d. samples from $\mathbb{D}_{PT}$ to construct $\tilde{D}_{PT}$
            \State $\text{datasets} \gets \text{datasets} \cup \{ \tilde{D}_{PT} \}$
            %  Pretrain a shadow model \textit{without} the challenge point
            \State $ g \leftarrow \mathcal{T}(\tilde{D}_{PT})$
            %  Pretrain a shadow model \textit{with} the challenge point
            \State Sample $\tilde{D}_{FT}$ i.i.d. using query access to $\mathbb{D}_{FT}$ 
            \State $f \leftarrow \mathcal{T}(g, \: \tilde{D}_{FT})$ \Comment{Finetune $g$ on $\tilde{D}_{FT}$}
            \State $\text{models} \gets \text{models} \cup \{f \}$ 
        \EndFor
        \Return $\text{models}, \: \text{datasets}$
    \end{algorithmic}
\end{algorithm}

\begin{figure}[h!]
    \centering
    \includegraphics[width=0.45\textwidth]{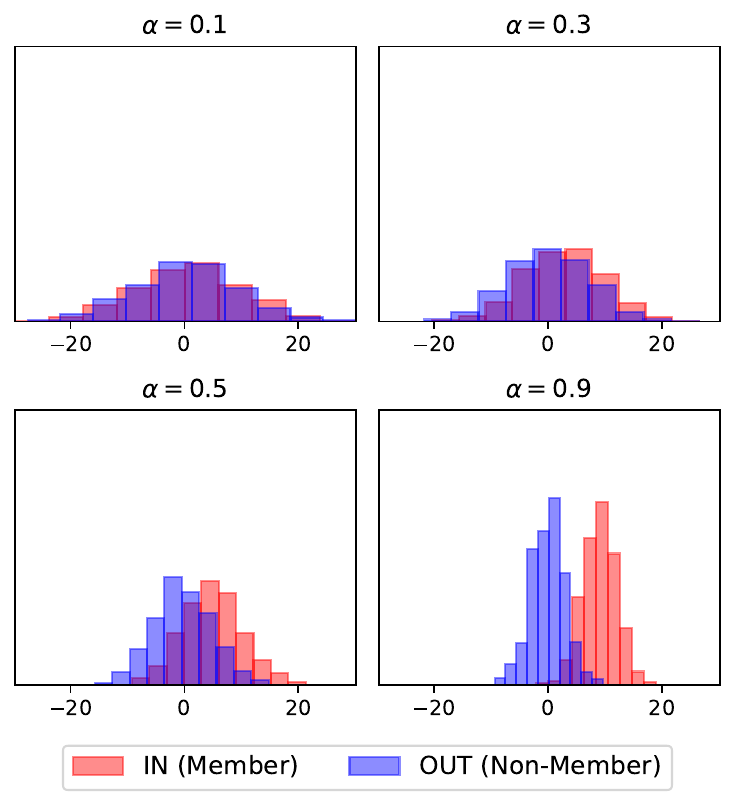}
    \caption{\revision{Distribution of the Test Statistic, $z$, for Multiple Values of $\alpha$}}
    \label{fig:theo_logits}
\end{figure}

\revision{
\subsection{Membership Inference Under Distribution Shift} \label{sec:dist_shift}

To motivate a membership-inference attack on finetuned deep learning models, we will first consider the simplified setting of mean estimation. A more detailed explanation, along with the proofs for the statements in this section, can be found in Appendix~\ref{adx:theoretical}.

Consider two datasets, $X \overset{\mathrm{iid}}{\sim} \mathcal{N}(\mu, \: \mathbb{I}_d)$ and $Y \overset{\mathrm{iid}}{\sim} \mathcal{N}(\mu + \nu, \: \mathbb{I}_d)$ where $\vert X \vert = n$, $\vert Y \vert = m$ such that $n \gg m$, and $\nu$ is a parameter that controls distribution shift. In this setting, the means of $X$ and $Y$ are related, and we would like to estimate the mean of $Y$, which has limited data, using the additional data from $X$. We define the estimator of $\mu + \nu$ as a combination of the empirical means of $X$ and $Y$:

\[
\hat{\mu} = \alpha \Bar{x} + (1-\alpha) \Bar{y}
\]

\noindent where $\alpha \in [0,1]$ and $\Bar{x}, \: \Bar{y}$ are the empirical means of $X$ and $Y$, respectively. Note that $\hat{\mu}$ has expected value and covariance 

% Here, $\alpha$ can be thought of as analogous to the epochs spent on pretraining, as a fraction of the total number of training epochs. When pretraining a machine learning model, the final learned parameter is an average of the gradients from each of the $n$ training steps. Suppose the same parameter is finetuned on a smaller dataset for $m \ll n$ epochs. The resulting parameter would take the same form as $\hat{\mu}$ where $\alpha$ equals the fraction of pretraining epochs $\frac{n}{n + m}$.

\[
\ex{}{\hat{\mu}} = \mu + (1-\alpha) \nu \\ 
\]

\[
\cov{}{\hat{\mu}} = \Big( \frac{\alpha^2}{n} + \frac{(1-\alpha)^2}{m} \Big) \cdot \mathbb{I}_d = \Tilde{\alpha} \cdot \mathbb{I}_d
\]

Suppose the challenger from the security game detailed in Section~\ref{sec:threatmodel} releases the statistic $\hat{\mu}$ and we, as the adversary, would like to learn samples' membership statuses with respect to the auxiliary (pretraining) data, $X$. One possible way to do this would be the following: Assume the adversary knows $\ex{}{\hat{\mu}}$ and $\mu$. Then, for some challenge point, $c$ the adversary can compute the test statistic 
\[
z = \inn{\hat{\mu} - \ex{}{\hat{\mu}}}{c - \ex{}{c}}
\]

% \john{Cite robust tracing and explain why we use this test statistic (preserve the information about how the challenge point contributes to the mean)}

This specific choice of test statistic is motivated by prior work on membership-inference attacks on published statistics~\cite{dwork-2015-trace}. Subtracting the expectation of each term allows the adversary to observe whether the noise from computing $\hat{\mu}$ is correlated with the noise from sampling $c$. Thus, the test statistic $z$ is a real number that measures the correlation between the challenge point and the published statistic, $\hat{\mu}$. The adversary can then choose a threshold $\tau$ such that if $z > \tau$, they will predict that the challenge point was IN (i.e. $c \in X$). Else, the adversary will predict that the challenge point was OUT (i.e. $c \sim \mathcal{N}(\mu, \: \mathbb{I}_d)$)

We will now show the ability of our attack to determine the membership status of the challenge point $c$ as a function of the parameter $\alpha$. To this end, we start by computing the expectation and variance of the test statistic, $z$, when $c$ is either OUT or IN.

\noindent \begin{lem} 
    If $c$ is OUT, then 
    \[
        \ex{}{z} = 0 \quad \textrm{and} \quad \var{}{z} = d \Tilde{\alpha},
    \] 
    and if $c$ is IN, then 
    \[
        \ex{}{z} = \frac{\alpha d}{n} \quad \textrm{and} \quad \var{}{z} = d \Tilde{\alpha} + \frac{2d \alpha^2}{n^2}
    \]
\end{lem} \label{lem:ex_var}

This lemma tells us that as long as the noise scale doesn't exceed the difference in means, it is straightforward to determine whether $c$ is IN or OUT. When $\alpha \to 0$, the published statistic is no longer encoding any information about $X$. Thus, the noise completely masks the difference in means, as shown in Figure~\ref{fig:theo_logits}. Conversely, as $\alpha \to 1$, we observe higher separation between the distributions of IN and OUT test statistics.

}
% \john{Can we explain the expectation/standard deviation ratio here?}

\revision{
Using Lemma~\ref{lem:ex_var}, we can analyze the performance (AUC) of the adversary's distinguishing test as a function of the parameter $\alpha$. To do this, we use the fact that the AUC of a classifier is equal to the probability that the classifier's prediction on a randomly chosen positive (IN) sample is greater than the prediction on a randomly chosen negative (OUT) sample \cite{roc-analysis}. Here, we use the assumption that the test statistic is normal. Because $z$ is the inner product of two high dimensional Gaussian vectors, and thus the sum of many i.i.d. Gaussian random variables, as $d \to \infty$, $z$ is normally distributed. 

\begin{lem}
    Assume that the test statistic, $z$, is normally distributed. Then, for a fixed $\alpha$, the AUC of our membership-inference attack can be written as
    \[
        AUC  = \frac{1}{2} \Bigg( 1 + \texttt{erf} \bigg(\frac{\alpha d}{2\sqrt{d ( \Tilde{\alpha}n^{2} + \alpha^{2}}) } \bigg) \Bigg) 
    \]
\end{lem} \label{lem:auc}

While it seems as if the attack's success is independent of the magnitude of the distribution shift, $\| \nu \|_2$, it is important to note that $\alpha$ should be set by the challenger such that the error on the new task (namely, estimating the mean of the new dataset, $Y$) is minimized. In this particular setting, $\alpha$ would be chosen to minimize the mean squared error between $\hat{\mu}$ and the mean of $Y$, $\mu + \nu$. The proof for the optimal setting of $\alpha$ can be found in Appendix~\ref{adx:results}. 
}

\begin{figure}[h!]
    \centering
    \includegraphics[width=0.55\textwidth]{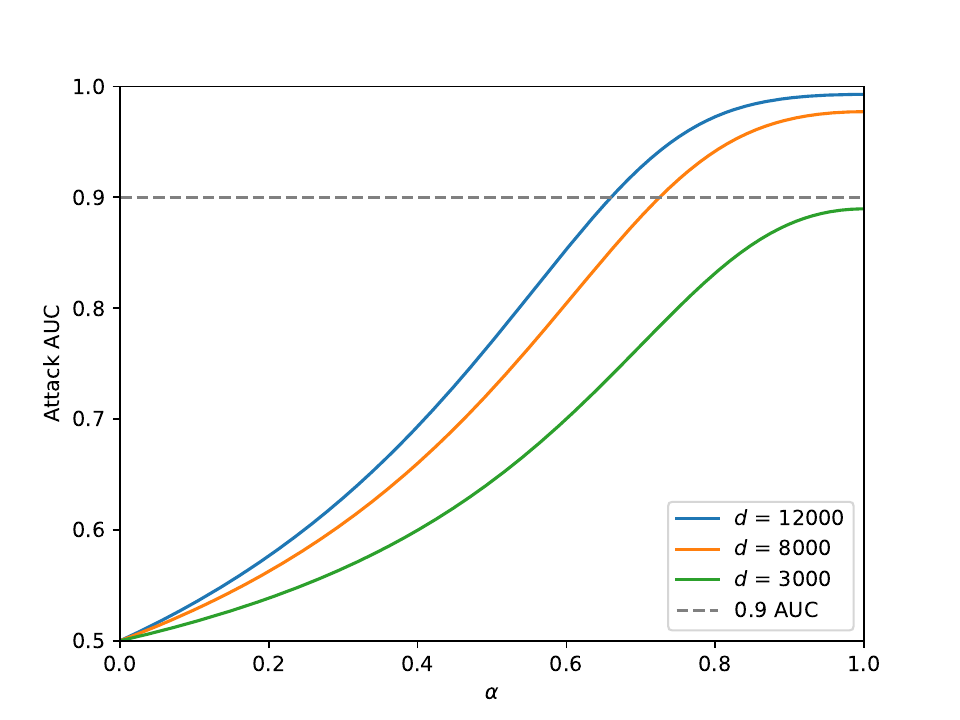}
    \caption{\revision{AUC of our Membership-Inference Attack on Mean Estimation as a Function of $\alpha$.}}
    \label{fig:AUC_theo}
\end{figure}

\revision{
Figure~\ref{fig:AUC_theo} visualizes the attack's AUC from Lemma~\ref{lem:auc} as a function of the parameter $\alpha$. Here, the parameters $n$, $m$, and $d$ are fixed. Our choices of the data's dimension, $d$, are motivated by the dimension of the data in our evaluation on vision models (Section~\ref{sec:experiments}). We observe that there is a rapid increase in AUC as more information about $X$ is preserved in the public statistic (i.e. as $\alpha$ increases). In this figure, the value for $\alpha$ when $d=12,000$ and $\| \nu \|_2 = 5$ that minimizes the error on the new task, $Y$, is roughly 0.77. This $\alpha$ corresponds to an AUC of 0.96. If we make the distribution shift larger, say $\| \nu \|_2 = 10$, the optimal value of $\alpha$ is 0.51, which corresponds to an AUC of 0.78. This shows that the success of our attack on mean estimation depends on the extent to which we combine the means of $X$ and $Y$ using the parameter $\alpha$, which is based on the similarity of the "pretraining" data $X$ and the "finetuning" data $Y$.
}

% Our findings for mean estimation are:

% \begin{itemize}
% \item As distribution shift increases (alpha is larger), ref to Fig 2 to say that separation is higher and the attack success is larger (Fig 3).
% \item How AUC varies with data dimension, this confirms results from literature \cite{}. As $d$ increases, the OUT test statistic is closer to 0, as the vectors will be orthogonal, but IN statistic is correlated with $\hat{\mu}$.
% \item Optimal choice of $\alpha$ (min loss) depends on distribution shift. 
% \item Discuss how this problem relates to finetuning (mean estimation = gradient descent; $\alpha$ is analogous to the number of pretraining epochs/total epochs).
% \end{itemize}

\subsection{Adapting an Existing Attack} 

As a first attempt to create an effective membership-inference attack on finetuned machine learning models, we can consider an adaptation of the \textit{likelihood ratio attack} (LiRA) proposed by Carlini et al. \cite{carlini-2022-lira}. In this attack (Algorithm~\ref{alg:adapted_lira}), the adversary observes the target model's prediction confidence on a challenge point with respect to the true label of the challenge point. Because the model's confidence with respect to a given label is approximately normally distributed, Carlini et al. perform a likelihood ratio test to infer the challenge point's membership status, using a set of shadow models to parameterize the IN and OUT distributions. 

In our setting, these shadow models are first trained on datasets drawn from $\mathbb{D}_{PT}$, then finetuned on a dataset drawn from $\mathbb{D}_{FT}$ (Algorithm~\ref{alg:sm_training}). Because the ground truth label of the challenge point drawn from $\mathbb{D}_{PT}$ is not necessarily in the range of our finetuned target model we cannot perform the likelihood ratio test with respect to the observed confidence on the point's true label. Instead, we can adapt the attack to use the label predicted by the target model with the highest confidence, $\hat{y}$. To do this, we  store the entire prediction vector for each query to our shadow models, and only use the scaled model confidences at index $\hat{y}$, denoted $f(x)_{\hat{y}}$, of the prediction vectors. We follow the lead of Carlini et al. \cite{carlini-2022-lira} and query each shadow and target model on $M$ random augmentations of the challenge point and fit $M$-dimensional multivariate normal distributions to the scaled model confidences we aggregate to improve attack success.

\begin{algorithm}
\caption{\textbf{Adapted LiRA} \\ We adapt the MI attack shown in \cite{carlini-2022-lira} by using the label which the target model predicted most confidently instead of the ground truth label. }\label{alg:adapted_lira}
    \begin{algorithmic}[1]
        \Require A finetuned target model $f_{\beta}$, a challenge point $\nolinebreak{x \leftarrow \mathbb{D}_{PT}}$, and models and datasets (i.e. the output of $\texttt{train\_shadow\_models}()$  )
        \State $\text{preds}_{\text{in}} \gets \{\}$, $\text{preds}_{\text{out}} \gets \{\}$
        \State $\vec{v}_{\text{obs}} \gets f_{\beta}(x)$ \Comment{Query the target model on $x$}
        \State $\text{conf}_{\text{obs}} \gets \texttt{logit}(\max_{i} \vec{v}_{\text{obs}, i})$ \Comment{Store max confidence score}
        \State $\hat{y} \gets \argmax_{i} \vec{v}_{\text{obs}, i}$ \Comment{Store most confident predicted label}
        
        % \State $\text{models}_{\text{in}} \gets \{\}$, $\text{models}_{\text{out}} \gets \{\}$ 
            % \State $f_{\text{in}} \gets \texttt{train\_shadow\_model}(x, 1)$ 
            % \State $\text{models}_{\text{in}} \gets \text{models}_{\text{in}} \cup \{ f_{\text{in}} \}$ 
            % \State $\text{models}_{\text{out}} \gets \text{models}_{\text{out}} \cup \{ f_{\text{out}} \}$
        \State $i \gets 1$ \Comment{Index for saved shadow models and datasets}
        \For{$N$ times} 
            \If{$x \in \text{datasets}_{i}$} \Comment{If $x$ is IN w.r.t. shadow model $i$}
                \State $f_{\text{in}} \gets \text{models}_{i}$
                \State $\text{conf}_{\text{in}} \gets \texttt{logit}(f_{\text{in}}(x)_{\hat{y}})$ \Comment{Query  $f_{\text{in}}$ on $x$}
                \State $\text{preds}_{\text{in}} \gets \text{preds}_{\text{in}} \cup \{ \text{conf}_{\text{in}} \}$ \Comment{Aggregate confidences}
            \ElsIf{$x \notin \text{datasets}_{i}$}
            %  Query $f_{\text{out}}$ on the challenge point to obtain the scaled confidence at label $\hat{y}$
                \State $f_{\text{out}} \gets \text{models}_{i}$
                \State $\text{conf}_{\text{out}} \gets \texttt{logit}(f_{\text{out}}(x)_{\hat{y}})$ 
                \State $\text{preds}_{\text{out}} \gets \text{preds}_{\text{out}} \cup \{ \text{conf}_{\text{out}} \}$
            \EndIf
            % \State $i \gets i + 1$
        \EndFor
        \State  $\mu_{\text{in}} \gets \texttt{mean}(\text{preds}_{\text{in}})$, \: $\mu_{\text{out}} \gets \texttt{mean}(\text{preds}_{\text{out}})$
        
        \State $\sigma^{2}_{\text{in}} \gets \texttt{var}(\text{preds}_{\text{in}})$, \: $\sigma^{2}_{\text{out}} \gets \texttt{var}(\text{preds}_{\text{out}})$        
        \State \textbf{return} $\dfrac{p(\text{conf}_{\text{obs}} \vert \mathcal{N}(\mu_{\text{in}}, \sigma^{2}_{\text{in}}))}{p(\text{conf}_{\text{obs}} \vert \mathcal{N}(\mu_{\text{out}}, \sigma^{2}_{\text{out}}))}$
    \end{algorithmic}
\end{algorithm}

\subsection{Issues with Adapting LiRA}

\begin{figure}[h!]
    \centering
    \includegraphics[width=0.5\textwidth]{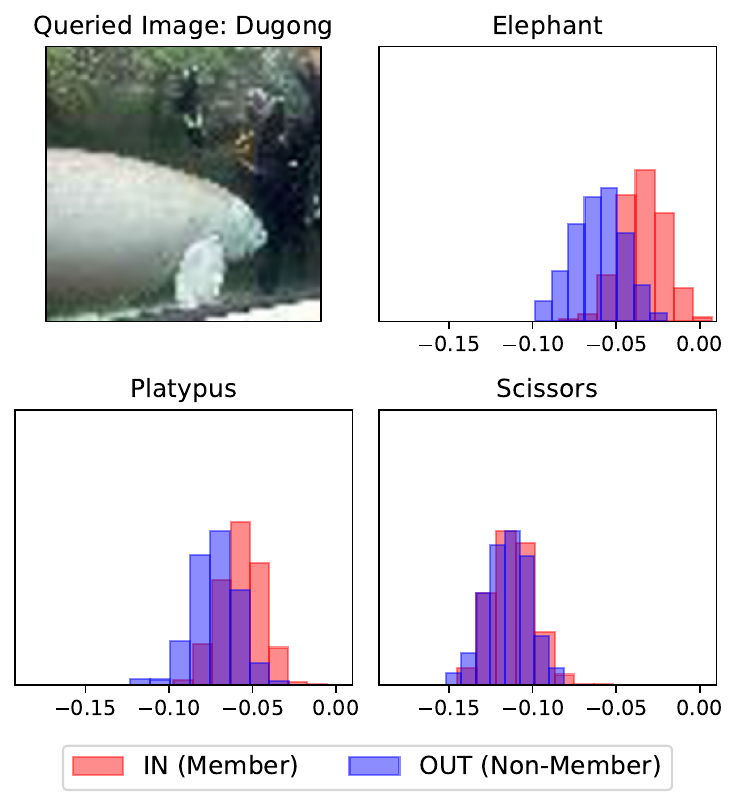}
    \caption{\revision{Scaled Model Confidences of Shadow Models Finetuned on Caltech 101 at Multiple Labels when Queried on a Sample from the "Dugong" Class in Tiny Imagenet}}
    \label{fig:multiple_logits}
\end{figure}

While this adaptation of LiRA is somewhat effective at inferring membership (Figures \ref{fig:feature_extraction-c100} and \ref{fig:feature_extraction-tiny}), it only captures how the pretraining dataset influences model's predictions with respect to a single label in the downstream dataset. Because the purpose of pretraining is to extract and learn general features that can be used in several downstream tasks, one would expect that the weights of a pretrained model have some impact on \textit{all} of a finetuned model's prediction confidences. \revision{For example, Figure~\ref{fig:multiple_logits} shows that the presence of a specific image labeled as "dugong" in the training set makes finetuned models, which cannot themselves predict the label "dugong", more confident on their downstream prediction of "elephant" and "platypus". Meanwhile, the presence of this image in the training dataset has little to no impact on the downstream label "scissors".}

Furthermore, if we observe the distribution of scaled model confidences over our shadow models, we see that it is approximately normal regardless of the choice of label. This may lead one to believe that the correct adaptation of LiRA to our setting would be to fit a multivariate normal distribution to the entire prediction vectors output by our shadow models. The assumption that the adversary only receives model confidences interferes with this seemingly better adaptation because of the softmax activation function. When softmax is applied, it converts the logit vector $\vec{z}$ into a probability distribution, $\vec{y}$, over the labels. Thus, the entries of $\vec{y}$ can be written as 

% \begin{align*}
% \vec{y} =  \: \begin{pmatrix}
%                                 p_{1} \\
%                                 p_{2} \\
%                                 \vdots \\
%                                 p_{K}
%                               \end{pmatrix} \in (0,1)^{K} 
% \end{align*}

\[
\vec{y} = (p_{1}, p_{2}, \dots, p_{K}) \in (0,1)^{K} 
\]

\noindent where $K$ is the number of classes and each $p_{i}$ denotes the model's confidence on class $i$. Because the entries of $\vec{y}$ necessarily sum up to $1$, any entry $p_{i}$ can be written as $1 - \sum_{j \neq i}{p_{j}}$. When we scale model confidences to compute the individual logits, $z_{i}$, any given computed logit can be written as a combination of the others. This means that our computed logits actually lie on a $(K-1)$-dimensional subspace of the $K$-dimensional space where the model's actual logits lie, and we cannot fit a $K$-dimensional multivariate normal distribution to all of our models' logit scaled prediction vectors without arbitrarily removing one of the entries in $\vec{y}$.

% This means that we cannot assume that $\mathbf{z}$ is distributed as a multivariate normal, since any entry of $\mathbf{z}$ is dependent on all of the other entries.

% \myparagraph{Attack Intuition.} Our attack uses shadow models to approximate the distribution of the target model's predictions with respect to a given challenge point. In a similar fashion to prior work, we train shadow models both with and without the challenge point to simulate the world where $b = 0$ and the world where $b = 1$. These shadow models are finetuned, using pretrained weights, on a task with different labels from the pretraining task, so we cannot use techniques that rely on knowing the challenge point's true label \cite{LiRA} to perform a distinguishing test. Therefore, to capture the target model's prediction behavior on the challenge point, we incorporate information from predictions over all possible labels. 

\subsection{Our \attackname\ Attack}

\indent Our \textbf{T}ransfer \textbf{M}embership \textbf{I}nference (\attackname) attack (Algorithm \ref{alg:transfer_mi}) starts with the same shadow model training procedure as Algorithm~\ref{alg:adapted_lira}, where the adversary trains shadow models on datasets sampled from $\mathbb{D}_{PT}$ and finetunes them on datasets sampled from $\mathbb{D}_{FT}$. The adversary then queries the challenge point on these shadow models to construct a dataset, $D_{\text{meta}}$, comprised of logits attained from scaling the prediction vectors as described in Section~\ref{subsection:logit_scaling}. To construct a distinguishing test that circumvents the issues that arise when attempting to parameterize the distribution of prediction vectors, the adversary trains a \textit{metaclassifier} on a collection of labeled prediction vectors $D_{\text{meta}}$, queries the target model on the challenge point, and scales the target model's prediction vector. Lastly, the adversary queries this observed prediction vector on their metaclassifier, which outputs a score in the interval $[0,1]$ that indicates the predicted membership status of the challenge point. Using a metaclassifier attack, \attackname, is still able to leverage the influence of memorized pretraining samples on predictions in the downstream task while not having to arbitrarily discarding one of the entries from the prediction vector. 

% Talk about querying several augmentations of the challenge point

In our implementation of \attackname\ for computer vision models, we train a metaclassifier per challenge point. Because we use a relatively small number of shadow models (64 IN and 64 OUT in total), we leverage random augmentations to construct a larger metaclassifier dataset. Each time we query the target model or our local shadow models, we query $M$ times with different random augmentations of the challenge point, including random horizontal flips and random crops with padding. This yields $M \times 2 \times 64$ prediction vectors for each challenge point. In total, our metaclassifiers are trained on 1024 labeled prediction vectors, 512 labeled $0$ to denote "non-member" or OUT and 512 labeled $1$ to denote "member" or IN. 

Due to computational limitations, we do not pretrain any shadow models for our attacks in the language domain. Rather, we use a publicly hosted pretrained model and finetune it on a downstream task. Without control over pretraining, we cannot produce a metaclassifier dataset with prediction vectors from both IN and OUT shadow models with respect to a single challenge point. This scenario can be represented in Algorithm \ref{alg:sm_training} by ommiting lines 4, 5 and 6, where $g$ refers to the publicly hosted pretrained language model. As a result, we use a \textit{global metaclassifier}, trained on a dataset containing the prediction vectors of \textit{all} challenge points, to produce membership scores. 

        % \For{$N$ times} \Comment{Train $2N$ shadow models and store them}
        %     \State $f_{\text{out}} \gets \texttt{train\_shadow\_model}(x, 0)$ 
        %     \State $f_{\text{in}} \gets \texttt{train\_shadow\_model}(x, 1)$ 
        %     \State $\text{models}_{\text{in}} \gets \text{models}_{\text{in}} \cup \{ f_{\text{in}} \}$ 
        %     \State $\text{models}_{\text{out}} \gets \text{models}_{\text{out}} \cup \{ f_{\text{out}} \}$
        % \EndFor

\begin{algorithm}[t]
\caption{\textbf{\attackname\ Metaclassifier Attack} \\ We pretrain shadow models with and without the challenge point and finetune them using query access to $\mathbb{D}_{FT}$ to estimate the target model's prediction behavior. Using the prediction vectors of our shadow models on the challenge point, we generate a dataset to train a metaclassifier to determine the challenge point's membership status.}\label{alg:transfer_mi}
    \begin{algorithmic}[1]    
        \Require A finetuned target model $f_{\beta}$, a challenge point $\nolinebreak{x \leftarrow \mathbb{D}_{PT}}$, and models and datasets (i.e. the output of $\texttt{train\_shadow\_models}()$ )
        \State $\text{preds}_{\text{in}} \gets \{\}$, $\text{preds}_{\text{out}} \gets \{\}$
        \State $i \gets 1$ \Comment{Index for saved shadow models}
        \For{$N$ times} 
            \If{$x \in \text{datasets}_{i}$} \Comment{If $x$ is IN w.r.t. shadow model $i$}
                \State $f_{\text{in}} \gets \text{models}_{i}$
                \State $\vec{v}_{\text{in}} \gets \phi (f_{\text{in} }(x))$ \Comment{Query IN model on x}
                \State $\text{preds}_{\text{in}} \gets \text{preds}_{\text{in}} \cup \{ (\vec{v}_{\text{in}}, 1) \}$  \Comment{Store and label the}
                \Statex \hspace{55mm} \textcolor{teal}{prediction vector}
            \ElsIf{$x \notin \text{datasets}_{i}$}
            %  Query $f_{\text{out}}$ on the challenge point to obtain the scaled confidence at label $\hat{y}$
                \State $f_{\text{out}} \gets \text{models}_{i}$
                \State  $\vec{v}_{\text{out}} \gets \phi (f_{\text{out}}(x))$
                \State $\text{preds}_{\text{out}} \gets \text{preds}_{\text{out}} \cup \{ (\vec{v}_{\text{out}}, 0) \}$
            \EndIf
        \State $i \gets i + 1$ 
        \EndFor
        \State $D_{\text{meta}} = \text{preds}_{\text{in}} \cup \text{preds}_{\text{out}}$  \Comment{Construct the metaclassifier}
        \Statex \hspace{43mm} \textcolor{teal}{dataset}
        \State $\mathcal{M} \leftarrow \mathcal{T}(D_{\text{meta}})$ \Comment{Train a binary metaclassifier}
        \State $\vec{v}_{\text{obs}} = \phi (f_{\beta}(x))$ \Comment{Query the target model on $x$}
        \State Output $\mathcal{M}(\vec{v}_{\text{obs}})$
    \end{algorithmic}
    \label{alg:transfer_mi}
\end{algorithm}

\section{\attackname\ Evaluation}

We evaluate the performance of our \attackname\ attack on image models with \revision{two pretraining tasks and four} downstream tasks and public, pretrained language models with \revision{one pretraining task and two downstream tasks}. We evaluate the success of our attack as a function of the number of updated parameters, and we choose downstream tasks with differing similarity to the pretraining task to show how attack success depends on the relevance of the pretraining task to the downstream task.  Additionally, we observe the success of our attack when differential privacy \cite{dwork-2006-dmns} is used in the finetuning process, which is an increasingly popular technique to maintain utility while preserving the privacy of individuals in the dataset of downstream task~\cite{yu-2022-dpllm, bu-2023-dpbiasft, abadi-2016-dpsgd, he-2022-groupwisedp, ganesh-2023-pretraining, li-2022-llmstrongdp}. 

This section presents the results of our evaluation of \attackname\ and addresses the following research questions with respect to the datasets in our experiments:

\begin{enumerate}[label=\textnormal{(\arabic*)}]
    \item[\mylabel{rq:leakprivinfo}{\textbf{Q1}}:] Can finetuned models leak private information about their pretraining datasets via black-box queries?
    \item[\mylabel{rq:finetuning}{\textbf{Q2}}:] Does updating a model's pretrained parameters instead of freezing them prevent privacy leakage?
    \item[\mylabel{rq:similarity}{\textbf{Q3}}:] Does the similarity between the pretraining and downstream task affect the privacy risk of the pretraining set?
    \item[\mylabel{rq:languagemodels}{\textbf{Q4}}:] Can the attack be generalized to domains other than vision?
    \item[\mylabel{rq:publicmodels}{\textbf{Q5}}:] Is it feasible to mount our attack on finetuned models that are based on publicly hosted foundation models?
    \item[\mylabel{rq:privacy}{\textbf{Q6}}:] Is privacy leakage present even when a model is finetuned using differential privacy?
\end{enumerate}

\subsection{Datasets and Models}
% CIFAR100 for training base models 25k samples each
% Coarse, CIFAR10, and SVHN for finetuning
% SGD lr=0.01, cosine annealing, standard augmentations such as crop and flips
In this section we will discuss the datasets used in our evaluation of \attackname. We will also discuss our choices of pretraining and downstream tasks used in our evaluation. 

\revision{

\subsubsection{Datasets}

We pretrain our small vision models on CIFAR-100~\cite{CIFAR} and finetune them on a coarse-labeled version of CIFAR-100, CIFAR-10~\cite{CIFAR}, and Oxford-IIIT Pet~\cite{pets}. Our larger vision models are pretrained on Tiny ImageNet~\cite{tiny-imagenet} and finetuned on Caltech 101~\cite{caltech-101}. For our language tasks, we use publicly available pretrained WikiText-103~\cite{wikitext103} models and finetune them on DBpedia~\cite{dbpedia} and Yahoo Answers~\cite{2015-yahoo} topic classification datasets. A detailed description of the datasets used in our evaluation can be found in Appendix~\ref{sec:datasets}.

\subsubsection{Models}

For our vision tasks, we use the ResNet-34 \cite{ResNet} and Wide ResNet-101 \cite{wide-resnet} architectures. The ResNet  architecture has been widely used in various computer vision applications due to its superior performance and efficiency. ResNet is a convolutional neural network architecture that uses residual blocks, allowing it to effectively handle the complex features of images and perform well on large-scale datasets.

For our language tasks, we use the Transformer-XL \cite{transformerxl} model architecture. In particular, we use the pretrained Transformer-XL model from Hugging Face, which is trained on WikiText-103 \cite{wikitext103}, as our initialization for the downstream tasks. We finetune our pretrained language model architectures on the DBpedia ontology classification and Yahoo Answers topic classification datasets.}

\subsubsection{Shadow Model Training} \hfill

Here, we describe the shadow model training procedure for our vision tasks, which comprise the majority of our experiments. The details for how we train shadow models for our language task can be found in Section~\ref{sec:language_models}. A full description can be found in Appendix~\ref{adx:sm_training}

\revision{Our shadow model training involves two phases: pretraining and finetuning. In the pretraining phase, we train 129 models are trained on random 50\% splits of CIFAR-100 and Tiny ImageNet using SGD with weight decay and cosine annealing for 100 epochs (ResNet-34) or 200 epochs (Wide ResNet-101). Standard data augmentations are applied during training and querying. In the second phase, the shadow models have a subset of their weights frozen and their classification layer swapped to match new task. Then, they are finetuned on random subsets of downstream task datasets. During pretraining, we designate a random set of challenge points to evaluate the \attackname\ attack. Because we train on 50\% splits of the pretraining data, approximately half of the challenge points are IN and OUT for each shadow model. In each experiment, we select a shadow model to be the target and use the remaining 128 to mount our attack, yielding a total of 128 trials.}

\subsection{Metrics}

To evaluate the performance of \attackname, we use a set of metrics that are commonly used in the literature. The first metric is \textit{balanced attack accuracy}, which measures the percentage of samples for which our attack correctly identifies membership status. Although balanced accuracy is a common metric used to evaluate MI attacks~\cite{shokri-2016-mi, choquettechoo-2021-labelonly, sablayrolles-2019-whitebox, 2018-yeom-mioverfitting}, prior work~\cite{carlini-2022-lira} argues that it  is not sufficient by itself to measure the performance of MI attacks as privacy is not an average case metric~\cite{steinke-2020-averageblog}. Therefore, we also evaluate our attack using the \textit{receiver operating characteristic} (ROC) curve. 

The ROC curve provides us with several additional metrics that we can use to evaluate the performance of \attackname. In our evaluation, we plot the ROC curve on a log-log scale to highlight the true positive rate (TPR) at low false positive rates (FPR), and we measure the area under the curve (AUC) as a summary statistic. Additionally, we report the TPR at low, fixed FPR of 0.1\% and 1\%. These metrics give us a more complete picture of how well \attackname\ performs in different scenarios. 
% ROC curve gives us a few more metrics that we can use
% We plot ROC on a loglog scale
% TPR @ 0.1% and 1% TPR
% AUC
% and \textit{true positive rates} (TPR) at low, fixed false positive rates (FPR).  

\subsection{Experimental Results} \label{sec:experiments}

In this section, we will discuss the performance of our attack on a variety of target models with different finetuning streategies. We consider models finetuned using feature extraction, models finetuned by updating pretrained weights, models finetuned with differential privacy, and publicly hosted pretrained models.

During training, \revision{we designate 1000 and 2000 samples to be challenge points for CIFAR-100 and Tiny ImageNet, respectively}, and we run our attack for each of these challenge points on 128 different target models. We compare our results to performing LiRA \cite{carlini-2022-lira} directly on the pretrained model (i.e., the adversary has access to the model before it was finetuned) to provide an upper bound on our attack's performance.

\subsubsection{Feature Extraction} \hfill \label{sec:feature_extraction}

\noindent \textbf{\ref{rq:leakprivinfo}: Can finetuned models leak private information about their pretraining datasets via black-box queries?}

\revision{To answer this research question, we evaluate the success of our \attackname\ attack on models finetuned without updating any of the pretrained parameters (i.e. feature extraction). We consider three tasks in our experiments where feature extraction is used to finetune our target model: Coarse CIFAR-100, CIFAR-10, and Caltech 101. Because feature extraction relies on the pretrained model being relevant to the downstream task, we choose the two most similar downstream tasks to pretraining. Our attack's success depends on the target model having high utility on its respective task, so it is important to ensure that we choose downstream tasks that are similar or relevant to the pretraining task when using feature extraction to finetune models. To transfer the pretrained CIFAR-100 models to Coarse CIFAR-100 and CIFAR-10 and the pretrained Tiny ImageNet models to Caltech 101, we remove the final classification layer, and replace it with a randomly initialized classification layer which has proper number of classes for the new downstream task. The remaining weights are kept frozen throughout training.}
% Additionally, in Figure~\ref{fig:feature_extraction} at higher FPR (about 5\%), the TPR of \attackname\ is approximately equal to that of the LiRA membership-inference attack directly on the pretrained model.

\begin{figure*}[h!]
    \centering
    \begin{subfigure}[t]{.49 \linewidth}
      \includegraphics[width=\linewidth]{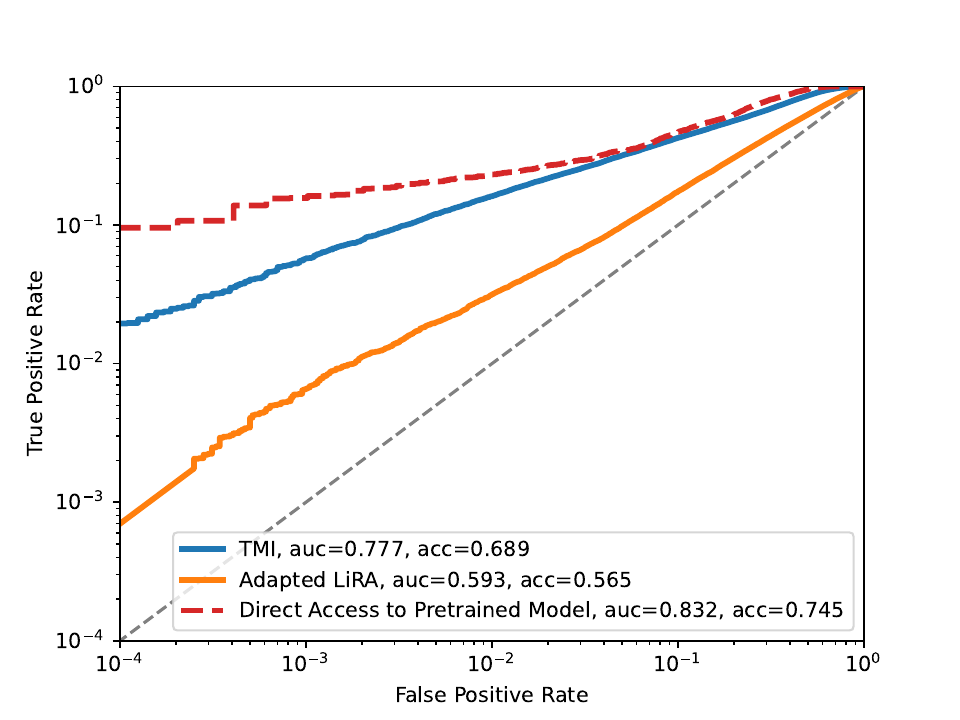}
      \caption{Coarse CIFAR-100}
    \end{subfigure}
    \begin{subfigure}[t]{.49\linewidth}
      \includegraphics[width=\linewidth]{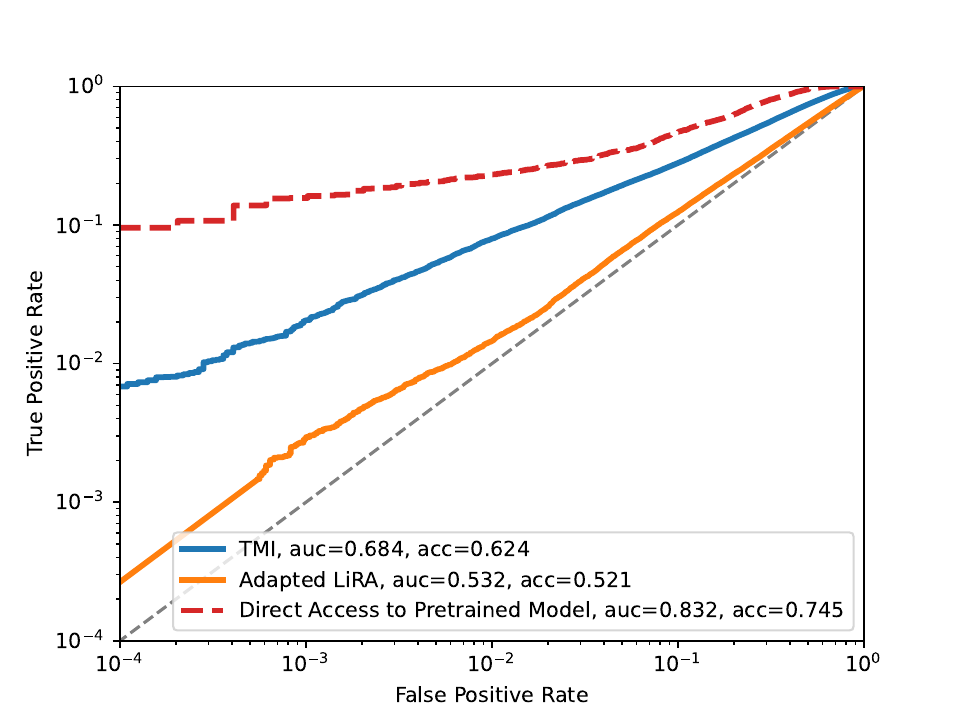}
      \subcaption{CIFAR-10}
    \end{subfigure}

    \caption{\attackname\ Attack Performance on Downstream Tasks When Preterained CIFAR-100 Target Models are Finetuned Using Feature Extraction}
    \label{fig:feature_extraction-c100}
\end{figure*}

\revision{As shown in Figure~\ref{fig:feature_extraction-c100}, we observe that \attackname\ is able to achieve AUC and balanced accuracy (0.78 and 69\%) within 0.06 of the adversary which has access to the pretrained model (0.83 and 75\%) on the Coarse CIFAR-100 downstream task. On this task, \attackname\ also has a TPR of 5.7\% and 16.1\% at 0.1\% and 1\% FPR, respectively. Despite being constrained to only having query access to the finetuned model, Figure~\ref{fig:feature_extraction-c100} shows that the TPR of \attackname\ is approximately equal at higher FPR (about 5\%) to that of running LiRA directly on the pretrained model.

Furthermore, Table~\ref{tab:tpr-feature-extraction} also shows the performance of \attackname\ on target models finetuned on the CIFAR-10 and Caltech 101 downstream tasks. When we run \attackname\ on the Tiny ImageNet models which are finetuned on Caltech 101, our attack achieves an AUC of 0.914, which is within ~6\% of the AUC achieved by LiRA directly on the pretrained model. As shown in Table~\ref{tab:tpr-feature-extraction}, \attackname\ has a $207\times$ and $41\times$ higher TPR than FPR when the FPR is fixed at 0.1\% and 1\%, respectively. On the CIFAR-10 finetuned models we observe that \attackname\ achieves a TPR of 2.0\% and 8.0\% at 0.1\% and 1\% FPR, respectively. Figure~\ref{fig:feature_extraction-c100} shows that \attackname\ also achieves an AUC of 0.684 and a balanced accuracy of 62.4\% when the downstream task is CIFAR-10. The lower attack success may be due to the relevance of the features learned during pretraining to the downstream task. For all three tasks, using our adaptation of LiRA and not incorporating information about all of the downstream labels yields \emph{significantly} lower performance by all of our metrics than \attackname. For example, at 0.1\% FPR, our attack has a TPR $14.7\times$, $8.1\times$, $6.7\times$ higher than adapted LiRA on Caltech 101, Coarse CIFAR-100 and CIFAR-10, respectively. \attackname\ also achieves an AUC about $1.3\times$ higher than adapted LiRA on the Coarse CIFAR-100 and CIFAR-10 tasks and an AUC $1.7\times$ higher on Caltech 101.}

% Report the accuracy of the models on each of the downstream tasks

\noindent \textbf{\ref{rq:leakprivinfo} Answer}: Yes, it is possible to infer the membership status of an individual in a machine learning model's pretraining set via query access to the finetuned model.

\begin{figure}[h!]
    \centering
    \includegraphics[width=0.5\textwidth]{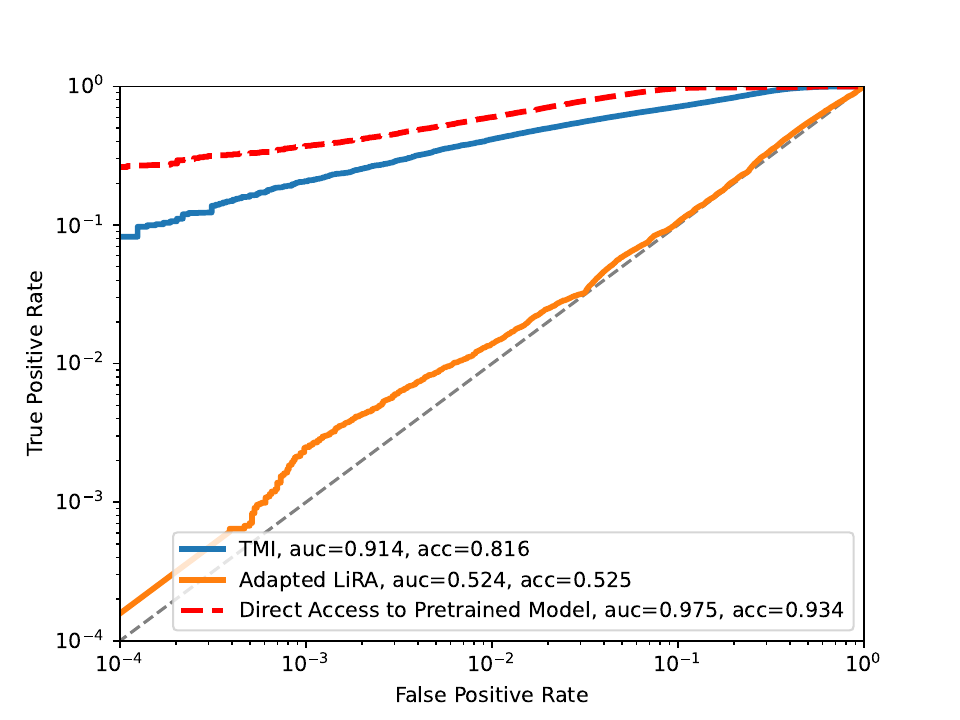}
    \caption{\revision{\attackname\ Attack Performance When Preterained Tiny ImageNet Target Models are Finetuned Using Feature Extraction}}
    \label{fig:feature_extraction-tiny}
\end{figure}

\begin{table}[h]
    \centering
    \caption{\revision{TPR at Fixed FPR of \attackname\ and Our Adaptation of LiRA when Pretrained Target Models are Finetuned Using Feature Extraction (Figures~\ref{fig:feature_extraction-c100} and \ref{fig:feature_extraction-tiny} )}}
    \label{tab:tpr-feature-extraction}
    
    \resizebox{.7\columnwidth}{!}{%
    \begin{tabular}{@{}lcc@{}}
    \toprule
        \multicolumn{1}{c}{\textbf{Task}}     & \textbf{TPR @ 0.1\% FPR} & \textbf{TPR @ 1\% FPR} \\ \midrule
        \attackname\ (CIFAR100 $\to$ Coarse CIFAR-100)  & 5.7\%  & 16.1\% \\
        \attackname\ (CIFAR100 $\to$ CIFAR-10)         & 2.0\%  & 8.0\%  \\
        \attackname\ (Tiny ImageNet $\to$ Caltech 101)      & 20.7\% & 41.5\% \\
        Adapted LiRA (CIFAR100 $\to$ Coarse CIFAR-100)         & 0.7\%  & 3.1\%  \\
        Adapted LiRA (CIFAR100 $\to$ CIFAR-10)                 & 0.3\%  & 1.5\%  \\
        Adapted LiRA (Tiny ImageNet $\to$ Caltech 101)      & 1.4\% & 0.25\% \\
        LiRA Directly on Pretrained Model (CIFAR-100)     & 15.6\% & 22.9\% \\ 
        LiRA Directly on Pretrained Model (Tiny ImageNet)     & 37.2\% & 60.1\% \\ \bottomrule \\ 
    \end{tabular}%
    }
\end{table}

\subsubsection{Updating Model Parameters} \hfill \label{sec:ft_model_parameters}

\noindent \textbf{\ref{rq:finetuning}: Does updating a model's pretrained parameters instead of freezing them prevent privacy leakage?}

% In this section, we will discuss the performance of our attack on target models that were transferred by updating a subset of the pretrained parameters. 

\paragraph{CIFAR-10} The ResNet models we pretrain on CIFAR-100 are divided into ResNet blocks or layers, which each contain multiple sub-layers. When finetuning pretrained ResNet models on CIFAR-10, we unfreeze the weights in different subsets of these ResNet layers. More concretely, we observe the performance of our attack on ResNet models which have had their classification layer (feature extraction), last 2 layers (62\% of total parameters), and last 3 layers (90\% of parameters) finetuned on the downstream task. 

\begin{figure}[]
    \centering
    \includegraphics[width=0.5\textwidth]{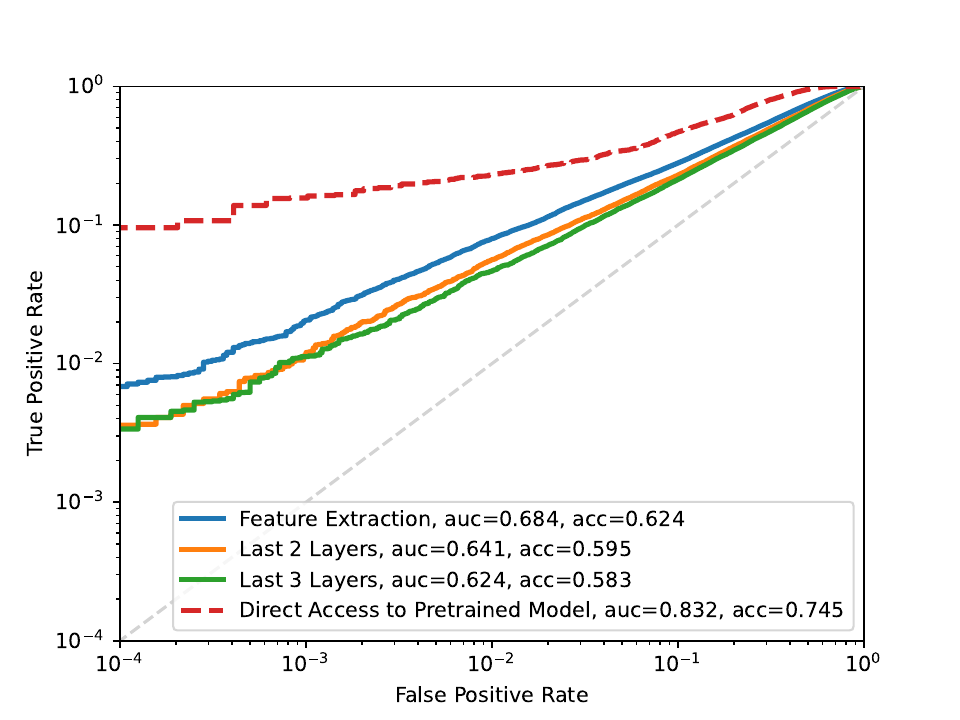}
    \caption{\attackname\ Performance when Finetuning Different Amounts of Parameters on CIFAR-10}
    \label{fig:ft_parameters}
\end{figure}

In Figure~\ref{fig:ft_parameters}, we observe that the AUC and accuracy of \attackname\ slightly decrease as we update an increasing number of parameters. We also observe a very slight decrease the TPR at a 1\% FPR when the number of finetuned parameters is increased from 2 layers to 3 layers, but TPR decreases at the FPR we consider when comparing to the TPR of \attackname\ on models finetuned with feature extraction. Table~\ref{tab:tpr-finetune} shows that updating the model's parameters induces a decrease in up to 0.8\% at a 0.1\% FPR and up to 3.3\% at a 1\% FPR.

\begin{figure}[h]
    \centering
    \includegraphics[width=0.5\textwidth]{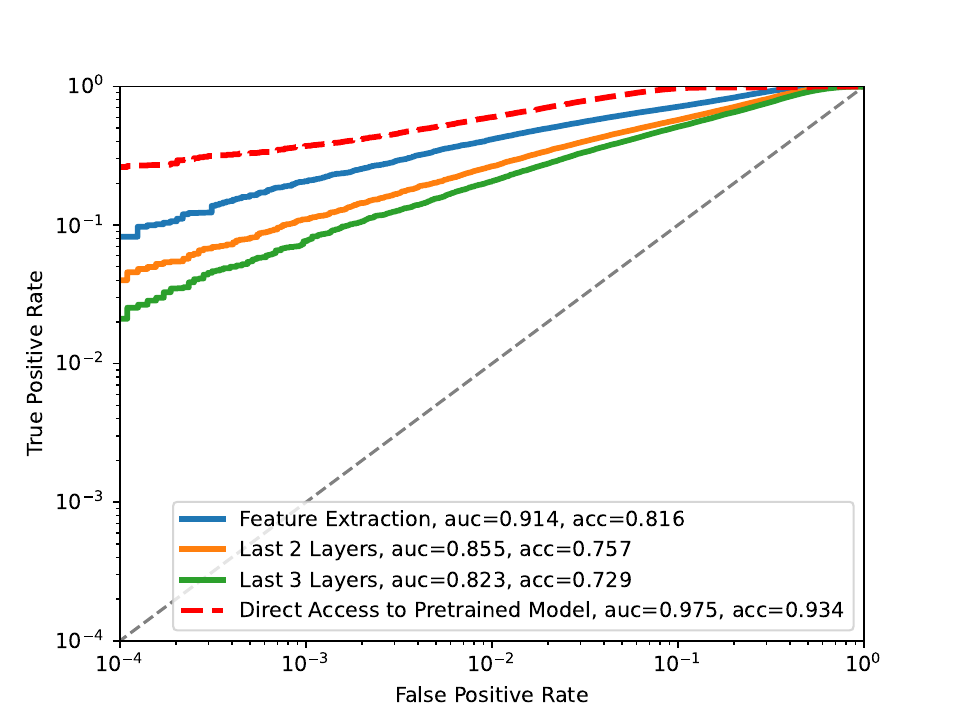}
    \caption{\revision{\attackname\ Performance when Finetuning Different Amounts of Parameters on Caltech 101}}
    \label{fig:ft_parameters_tiny}
\end{figure}

\revision{\paragraph{Caltech 101} The Wide ResNet models we pretrain on Tiny ImageNet have a similar architecure to the ResNets in the previous experiments, where each block contains sub-layers. For this architecture, we run our attack on models which have had their classification layer (feature extraction), last 2 layers (34\% of total parameters), and last 3 layers (96\% of parameters) finetuned on Caltech 101. In Figure~\ref{fig:ft_parameters_tiny},  we observe a similar trend to the previous experiments on CIFAR-10 models, where the attack's success decreases as we increase the number of finetuned parameters. In Table~\ref{tab:tpr-finetune} we see that for a fixed FPR of 0.1\%, \attackname\ has a 20.7\%, 11\%, and 7.7\% TPR when the final, last two, and last three layers are finetuned, respectively. At a 1\% FPR, \attackname\ has a 41.5\%, 26.5\% and 20.6\% TPR for these three settings. Nevertheless, \attackname\ achieves comparable AUC and balanced accuracy metrics to feature extraction when we finetune the majority of model parameters in both the CIFAR-10 and Caltech 101 experiments.
}

\revision{Prior work \cite{jagielski-2023-forgetting} has shown that samples used earlier in training are more robust to privacy attacks. Our theoretical results in Section~\ref{sec:dist_shift} substantiate this work and help provide an explanation for the decrease in attack success. In our results, $\alpha$ corresponds to the fraction of training epochs spent on finetuning, but our analysis lacks a critical parameter from our experiments: the number or fraction of tunable parameters in the published statistic. Our analysis considers a vector (namely, the empirical mean) where all of the parameters are being updated, thus providing a \textit{worst-case} situation for the adversary. In the feature extraction setting, the information learned by the model during pretraining is essentially frozen. Unlike feature extraction, we are updating the model's parameters with information about the downstream samples when finetuning.}

\noindent \textbf{\ref{rq:finetuning} Answer}: Updating larger subsets of model parameters slightly decreases the success of our \attackname\ attack when compared to models finetuned on downstream tasks using feature extraction, but we are still able to infer the membership status of the majority of samples in the pretraining dataset.

\begin{table}[h]
    \centering
    \caption{\revision{TPR at Fixed FPR of \attackname\ when Pretrained Target Models are Finetuned on by Updating the Pretrained Weights (Figures~\ref{fig:ft_parameters} and \ref{fig:ft_parameters_tiny})}}
    \label{tab:tpr-finetune}
    
    \resizebox{.7\columnwidth}{!}{%
    \begin{tabular}{@{}lcc@{}}
    \toprule
    \multicolumn{1}{c}{\textbf{Task}}     & \textbf{TPR @ 0.1\% FPR} & \textbf{TPR @ 1\% FPR} \\ \midrule
    Feature Extraction (CIFAR-100 $\to$ CIFAR-10)        & 2.0\%  & 8.0\%  \\
    Last 2 Layers (CIFAR-100 $\to$ CIFAR-10)           & 1.1\%  & 5.6\%  \\
    Last 3 Layers (CIFAR-100 $\to$ CIFAR-10)          & 1.1\%  & 4.7\%  \\ 
    
    Feature Extraction (Tiny ImageNet $\to$ Caltech 101)        & 20.7\%  & 41.5\%  \\
    Last 2 Layers (CIFAR-100 $\to$ CIFAR-10)           & 11.0\%  & 26.5\%  \\
    Last 3 Layers (CIFAR-100 $\to$ CIFAR-10)          & 7.7\%  & 20.6\%  \\ 
    
    \bottomrule \\ 
\end{tabular}%
}

\end{table}

\noindent \textbf{\ref{rq:similarity}: Does the similarity between the pretraining and downstream task affect the privacy risk of the pretraining set?}

\paragraph{Oxford-IIIT Pet} The Oxford-IIIT Pet dataset presents a unique challenge for finetuning our pretrained ResNet models. To finetune these models on the pet breeds classification task, it is necessary to unfreeze all of the layers. Otherwise, the model would have little to no utility with respect to the downstream task. Because the 37 pet breeds that appear in this dataset do not appear in and are not similar to any of the classes in the pretraining data, freezing any of the model's weights is an innefective strategy for this task. In this evaluation of \attackname\ on models transferred from CIFAR-100 to Oxford-IIIT Pet, we finetune for the same number of epochs with the same hyperparameters as the models in our experiments with CIFAR-10.

\begin{table}[h]
    \centering
    \caption{TPR at Fixed FPR of \attackname\ when Target Models are Finetuned on Oxford-IIIT Pet by Finetuning All Layers}
    \label{tab:tpr-pets}
    \resizebox{.55\columnwidth}{!}{%
    \begin{tabular}{@{}lcc@{}}
    \toprule
    \multicolumn{1}{c}{\textbf{Task}}     & \textbf{TPR @ 0.1\% FPR} & \textbf{TPR @ 1\% FPR} \\ \midrule
    \attackname\ (Oxford-IIIT Pet)    & 0.5\%  & 2.6\%   \\ 
    Adapted LiRA (Oxford-IIIT Pet)    & 0.08\% & 1.0\% \\ \bottomrule \\
    \end{tabular}%
    }
\end{table}

\begin{figure}[h]
    \centering
    \includegraphics[width=0.5\textwidth]{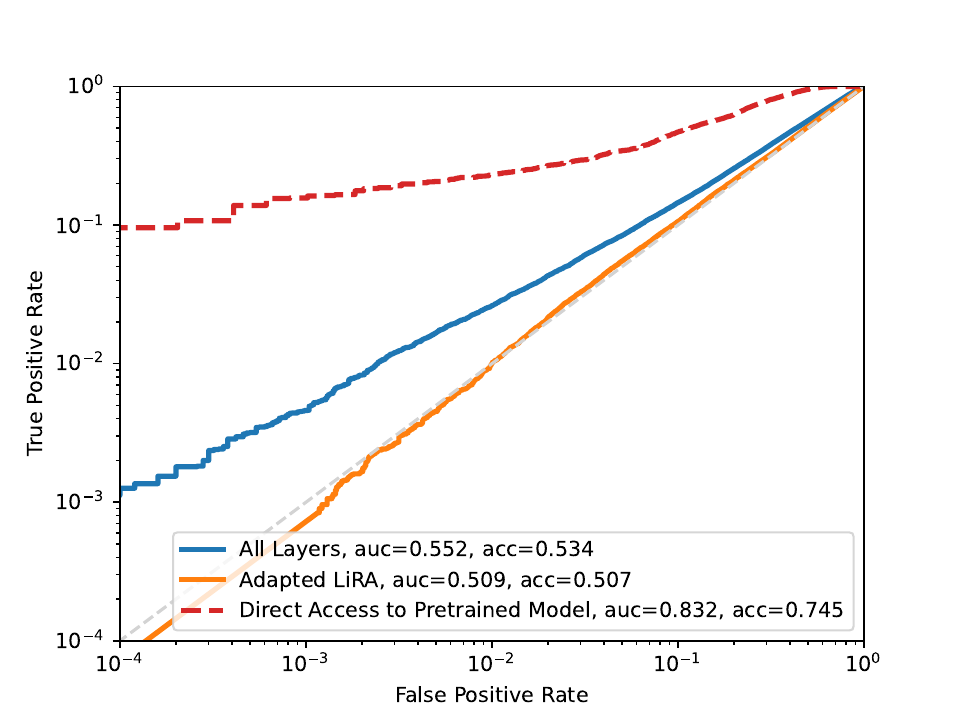}
    \caption{\attackname\ Performance when Finetuning All Parameters on the Oxford-IIIT Pet Dataset}
    \label{fig:ft_pets}
\end{figure}

We observe in Figure~\ref{fig:ft_pets} that the accuracy and AUC of our adaptation of LiRA becomes effectively as good as randomly guessing membership status. In contrast, \attackname\ is still able to achieve some amount of success, with an AUC of 0.55 and a balanced accuracy of 53.4\% over 128 target models with 1000 challenge points each. Additionally, our attack demonstrates a 2.6\% true positive rate at a 1\% false positive rate. 

\noindent \textbf{\ref{rq:similarity} Answer}: Even though the downstream task of pet breed classification is dissimilar from the pretraining task and all of the model's parameters are finetuned for 20 epochs, \attackname\ is able to achieve non-trivial success metrics when inferring the membership status of samples in the pretraining dataset.

% TranferMI with tasks that diverge from pretraining

\subsubsection{Finetuning Pretrained Language Models} \hfill \label{sec:language_models}

\noindent \textbf{\ref{rq:languagemodels}: Can the attack be generalized to domains other than vision?}

To answer this research question, we evaluate the success of our \attackname\ attack in the natural language domain. In particular, we focus on publicly available pretrained large language models (LLMs), or foundation models~\cite{foundationmodels}, which we finetune on \revision{two text classification tasks}. 

Due to computational limitations, we do not train LLMs from scratch. As an alternative, we evaluate our attack on a widely used pretrained foundation model, Transformer-XL~\cite{transformerxl}, along with its corresponding tokenizer, which are hosted by Hugging Face~\cite{huggingface}. Only a limited number of organizations with sufficient computational resources possess the capability to train foundation models, which are typically fine-tuned on specific tasks by smaller organizations \cite{openai-api, google-api, microsoft-api}. Through our evaluation of \attackname\ on finetuned foundation models, we will additionally answer the following research question:

\noindent \textbf{\ref{rq:publicmodels}: Is it feasible to mount our attack on finetuned models that are based on publicly hosted foundation models?}

We chose this foundation model in particular because it uses known training, validation, and testing splits from the WikiText-103~\cite{wikitext103} dataset, providing us with the exact partitions necessary to evaluate \attackname\ without having to train our own LLMs. Additionally, although modest in comparison to contemporary foundation models, the Transformer-XL architecture contains 283 million trainable parameters. This makes it a powerful and expressive language model that may be prone to memorizing individual data points.

We finetune Transformer-XL on DBpedia~\cite{dbpedia}, modifying the pretrained tokenizer to use a max length of 450, including both truncation and padding. Using a training set of 10,000 randomly sampled datapoints from DBpedia, we finetune the last third of the parameters in our Tranformer-XL models for 1 epoch. We use the AdamW~\cite{loshchilov-2017-adamw} optimizer with a learning rate of $10^{-5}$ and weight decay with $\lambda=10^{-5}$. With these hyperparameters, we are able to achieve test accuracies of $97\%$ \revision{and $60\%$} on the 14 classes of DBpedia \revision{and 10 classes of Yahoo Answers, respectively}.

To prepare our membership-inference evaluation dataset, the WikiText-103 is partitioned into contiguous blocks, separated each by Wikipedia subsections. We then perform the same tokenization process as we do in finetuning before collecting their prediction vectors. Because we do not pretrain our own LLMs, we adapt \attackname\ to train a single, global metaclassifier over the prediction vectors of all challenge points rather than train a metaclassifier per challenge point. In total, we use 2650 challenge points, which corresponds to a metaclassifier dataset with size $|D_{\text{meta}}| = 2560 * (\text{number of shadow models}$).

% \sw{do i need to explain this}

% Note that this is because we are unable to train $\vec{v}_{\text{in}}$ and $\vec{v}_{\text{out}}$ for any given target point like in the vision domain.
We are unable to compare \attackname\ to our adaptation of LiRA because we cannot pretrain our own LLMs. Our adaptation of LiRA requires additional shadow models to be trained from scratch with respect to every challenge point as detailed in Algorithm~\ref{alg:adapted_lira}. In our evaluation, we also find that k-nearest neighbors (KNN) significantly outperforms a neural network as a global metaclassifier. We believe this to be the case due to the additional variance incurred in a (global) metaclassifier dataset containing prediction vectors from all challenge points. In contrast, the metaclassifier datasets used in our vision tasks only contained labeled prediction vectors with respect to a single challenge point.

\begin{figure}[h]
    \centering
    \includegraphics[width=0.5\textwidth]{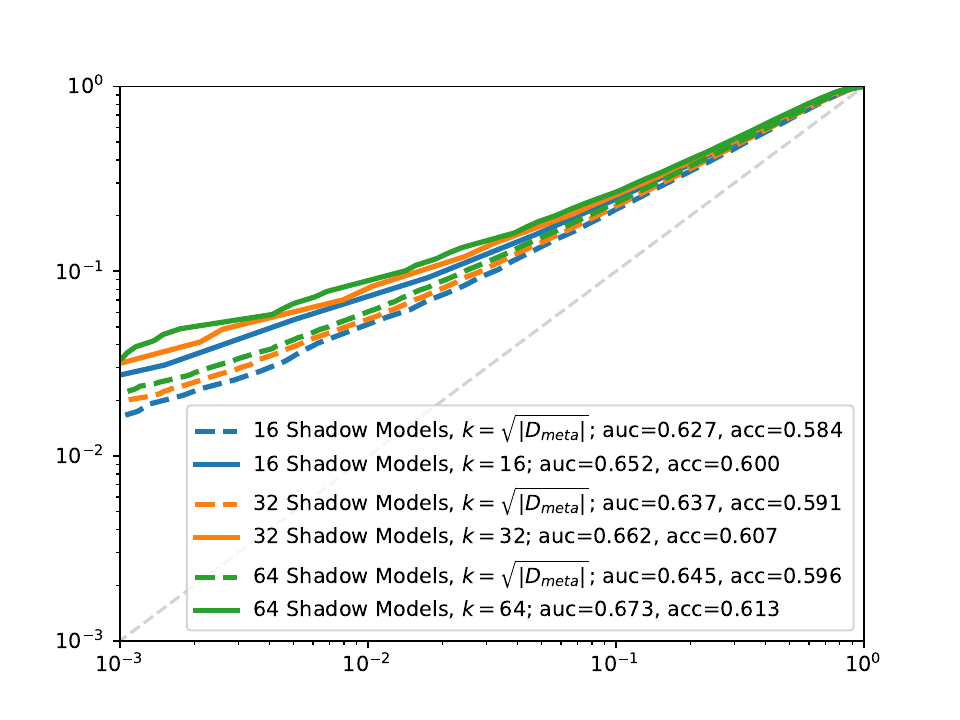}
    \caption{\attackname\ Performance on a Publicly Available Transformer-XL Model Finetuned on DBpedia-14 Topic Classification}
    \label{fig:language_dbpedia}
\end{figure}

\begin{figure}[h]
    \centering
    \includegraphics[width=0.5\textwidth]{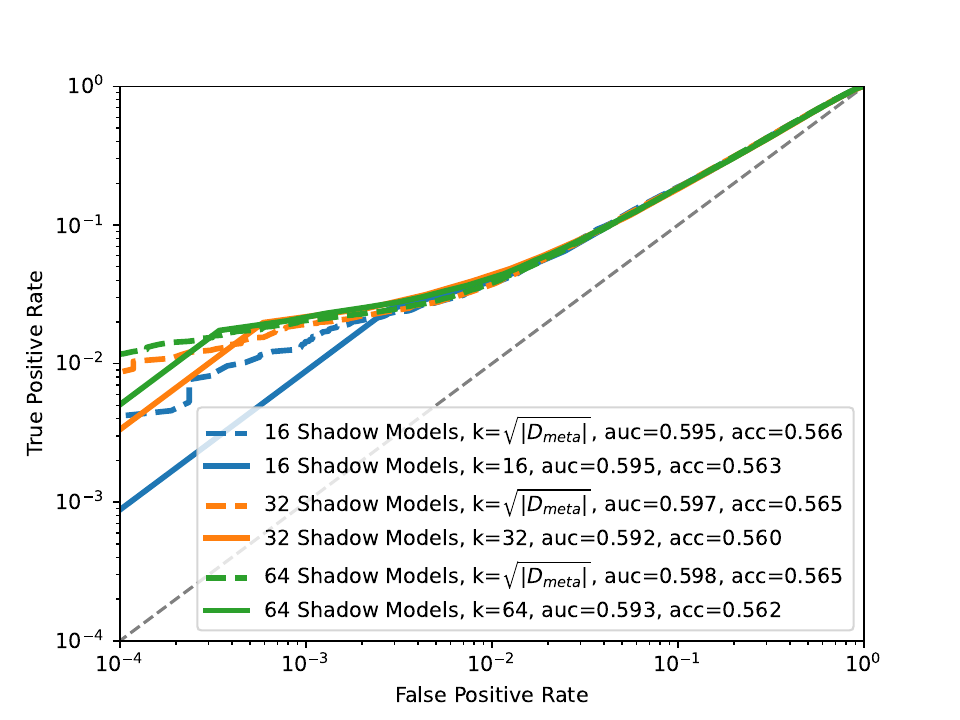}
    \caption{\revision{\attackname\ Performance on a Publicly Available Transformer-XL Model Finetuned on Yahoo Answers Topic Classification.}}
    \label{fig:language_yahoo}
\end{figure}

We present the results of our evaluation on LLMs in \revision{Figures~\ref{fig:language_dbpedia} and \ref{fig:language_yahoo} and Tables~\ref{tab:language} and \ref{tab:language_yahoo}}. Although it is common practice to use $k = \sqrt{n}$ neighbors in a KNN, we also report results using $k$ equal to the number of shadow models as it appears to increase attack success. As shown in Table~\ref{tab:language}, we observe that \attackname\, using the highest number of shadow models (64), is able to achieve a TPR of 3.4\% and 8.8\% at 0.1\% and 1\% FPR, respectively. These results are comparable to our findings on CIFAR-10 from Table~\ref{tab:tpr-finetune} in the vision domain. Surprisingly, we do not observe a notable difference in our summary statistics as we increase the number of shadow models from 16 to 64, with an increase of only 0.652 to 0.673 in AUC, and 60\% to 61.3\% in accuracy as shown in Figure~\ref{fig:language_dbpedia}.

\revision{The results shown in Figure~\ref{fig:language_yahoo} and Table~\ref{tab:language_yahoo} are consistent with our finding in Section~\ref{sec:ft_model_parameters} for similarity between the pretraining and downstream task. We see a slight decrease in \attackname's success when Transformer-XL is finetuned on a completely new task, Yahoo Answers, versus when it is finetuned on data from a similar distribution to its pretraining, DBpedia-14. When the Transformer-XL model is finetuned on the Yahoo Answers topic classification task, \attackname\ achieves a TPR of 2.6\% and 4.2\% at 0.1\% and 1\% FPR, respectively. Compared to our previous language model experiments, the \attackname\ sees a decrease in AUC from 0.67 to 0.59 and a slight decrease in accuracy from 61\% to 56\%.}

\begin{table}[h!]
\centering
\caption{TPR at Fixed FPR of \attackname\ on Pretrained WikiText-103 Transformer-XL Finetuned on DBpedia-14 (Figure~\ref{fig:language_dbpedia})}
\label{tab:language}
\resizebox{.6\columnwidth}{!}{%
\begin{tabular}{@{}lcc@{}}
    \toprule
    \multicolumn{1}{c}{\textbf{Task}}     & \textbf{TPR @ 0.1\% FPR} & \textbf{TPR @ 1\% FPR} \\ \midrule
    16 Shadow Models ($k=\sqrt{\vert D_{\text{meta}} \vert}$)  & 1.6\%  & 5.2\% \\
    16 Shadow Models ($k=16$) & 2.6\%  & 7.0\% \\
    32 Shadow Models ($k=\sqrt{\vert D_{\text{meta}} \vert}$)      & 2.0\%  & 5.5\% \\
    32 Shadow Models ($k=32$) & 3.1\%  & 8.1\% \\
    64 Shadow Models ($k=\sqrt{\vert D_{\text{meta}} \vert}$)      & 2.2\%  & 6.0\% \\
    64 Shadow Models ($k=64$)  & 3.4\%  & 8.8\% \\ \bottomrule \\ 
    \end{tabular}%
    }

\end{table}

\noindent \textbf{\ref{rq:languagemodels} Answer:} Yes, we are able to generalize \attackname\ to domains other than vision. In particular, we are able to show that our attack is effective against pretrained language models, and present our results on the publicly hosted Transformer-XL foundation model without the need to pretrain any additional large language models.

\noindent \textbf{\ref{rq:publicmodels} Answer:} Yes, \attackname\ continues to be effective in this situation where we finetuned public foundation models. This reinforces the need for understanding privacy leakage in the transfer learning setting used for foundation models.

% Q5: Is it feasible to perform our attack on finetuned public models?
% We adapt the attack as a proxy for running the original attack because of computational constraints
% 1. Talk about experimental (Justify datasets/models)
%   - Datasets, tokenizer, model arch, how much we finetune
%   -  We know the training set (evaluation), the model has 300M parameters, downstream/finetune task is similar to pretraining (97% accuracy)
%   - How we create challenge pts for wikitext103
% 2. Adapting the attack to use global metaclassifier
%   - How we adapted the attack, why we use a KNN metaclassifier 
% 3. Mention that we can't compare with LiRA w/ direct access because it's computationally infeasible to train any number of OUT models
% 4. Discuss the results 
%   - TPR @ low FPR, 

\begin{table}[h!]
\centering
\caption{\revision{TPR at Fixed FPR of \attackname\ on Pretrained WikiText-103 Transformer-XL Finetuned on Yahoo Answers (Figure~\ref{fig:language_yahoo})}}
\label{tab:language_yahoo}
\resizebox{.6\columnwidth}{!}{%
\begin{tabular}{@{}lcc@{}}
\toprule
\multicolumn{1}{c}{\textbf{Task}}     & \textbf{TPR @ 0.1\% FPR} & \textbf{TPR @ 1\% FPR} \\ \midrule
16 Shadow Models ($k=\sqrt{\vert D_{\text{meta}} \vert}$)  & 1.4\%  & 3.8\% \\
16 Shadow Models ($k=16$) & 1.1\%  & 4.0\% \\
32 Shadow Models ($k=\sqrt{\vert D_{\text{meta}} \vert}$)      & 1.9\%  & 3.7\% \\
32 Shadow Models ($k=32$) & 2.0\%  & 4.4\% \\
64 Shadow Models ($k=\sqrt{\vert D_{\text{meta}} \vert}$)      & 2.1\%  & 3.8\% \\
64 Shadow Models ($k=64$)  & 2.6\%  & 4.2\% \\ \bottomrule \\ 
\end{tabular}%
}

\end{table}

\subsubsection{Transfer Learning with Differential Privacy} \hfill \label{sec:differential_privacy}

\noindent \textbf{\ref{rq:privacy}: Is privacy leakage present even when a model is finetuned using differential privacy?}

We also discuss the performance of our attack on target models that were finetuned with differential privacy. Because prior work on transfer learning with differential privacy considers strategies where an especially small percentage of parameters are trained on the downstream task \cite{yu-2022-dpllm, bu-2023-dpbiasft, abadi-2016-dpsgd, papernot-2020-making}, we freeze the pretrained model's weights and train only the final layer on the downstream task. In our experiments, we perform feature extraction to finetune our pretrained CIFAR-100 models on Coarse CIFAR-100 and CIFAR-10. We train the final classification layer using DP-SGD \cite{abadi-2016-dpsgd} with target privacy parameters $\varepsilon = \{0.5, 1\}$ and $\delta = 10^{-5}$. As these are strict privacy parameters, we set the clipping norm equal to $5$ to achieve reasonable utility on the downstream tasks.

\begin{figure*}[h!]
\centering
    \begin{subfigure}[t]{.49\linewidth}
      \includegraphics[width=\linewidth]{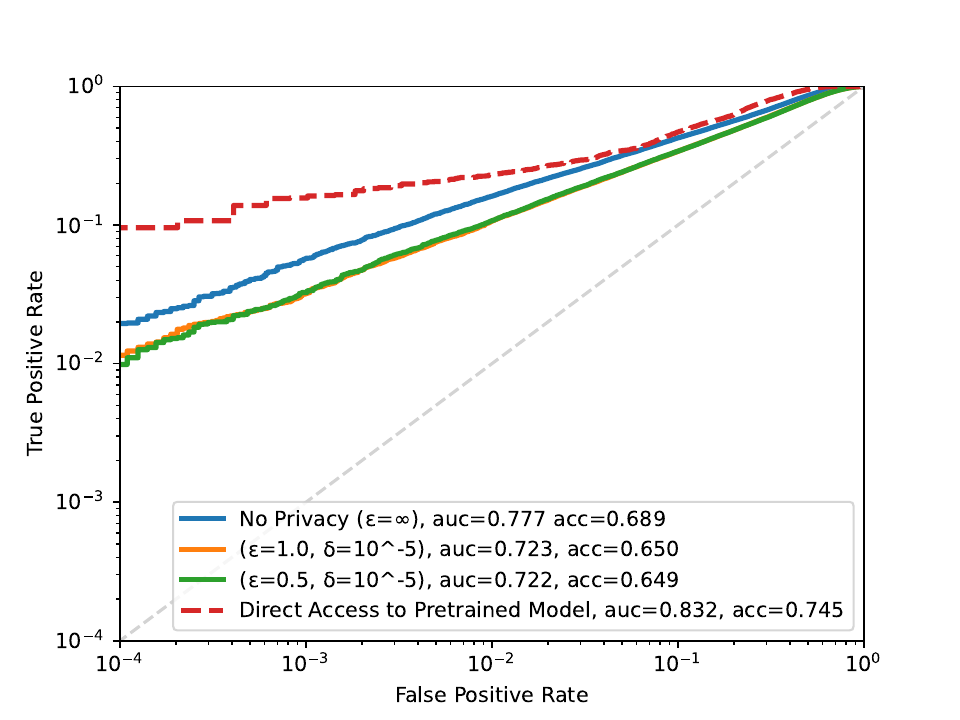}
      \caption{Coarse CIFAR-100}
    \end{subfigure}
    \begin{subfigure}[t]{.49\linewidth}
      \includegraphics[width=\linewidth]{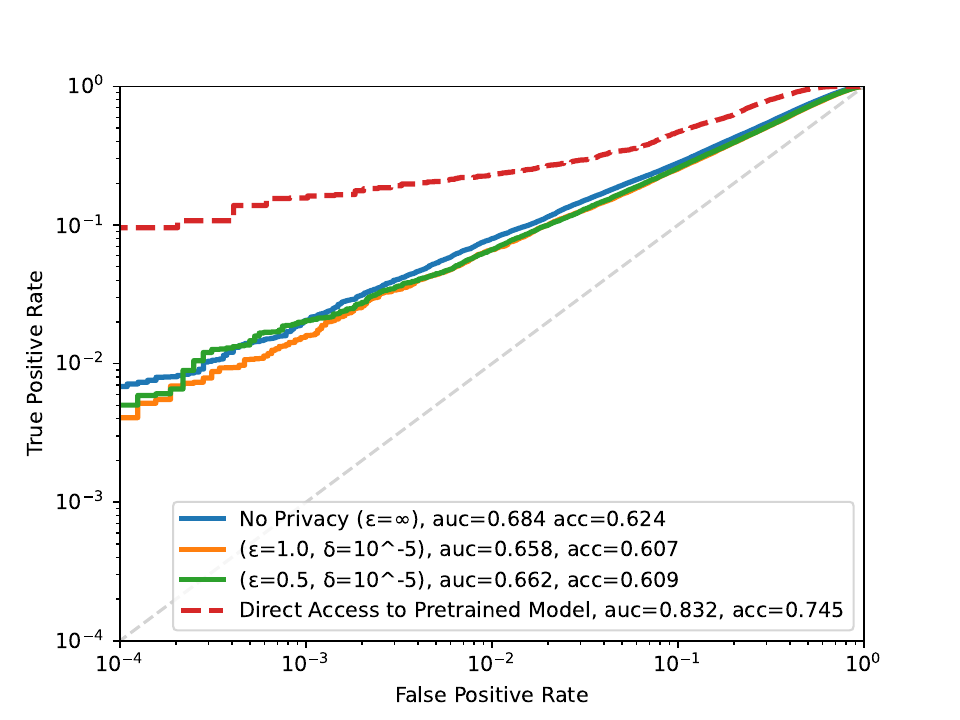}
      \subcaption{CIFAR-10}
    \end{subfigure}
    \captionsetup{justification=centering}
    \caption{\attackname\ Performance when Finetuning Models with Differential Privacy (DP-SGD)}
    \label{fig:dpsgd}
\end{figure*}

Figure~\ref{fig:dpsgd} shows that the success of our attack only decreases slightly when differential privacy is used to train the final classification layer on a downstream task. We believe that the slight decrease in attack accuracy can be attributed to loss in utility with respect to the downstream task from training with DP-SGD. When we finetune our models on Coarse CIFAR-100 with privacy parameters $\varepsilon=0.5$ and $\delta=10^{-5}$, \attackname\ has a TPR of 3.3\% at a FPR of 0.1\% and a TPR of 10.7\% at a FPR 1\%. Additionally, our attack maintains about 95\% of the accuracy and AUC compared to the setting where no privacy preserving techniques are used to finetune models on Coarse CIFAR-100.

Prior work \cite{carlini-2022-lira} has shown that state-of-the-art MI attacks, which directly query the pretrained model, completely fail when the target models are trained with a small amount of additive noise. For example, when training target models using DP-SGD with a clipping norm equal to $5$ and privacy parameter $\varepsilon=8$, LiRA has an AUC of 0.5. Through this evaluation, we reinforce the fact that transferring pretrained models to downstream tasks with differential privacy does not provide a privacy guarantee for the pretraining data.

\begin{figure*}[ht]
    \centering
    \begin{subfigure}[t]{.49\linewidth}
      \includegraphics[width=\linewidth]{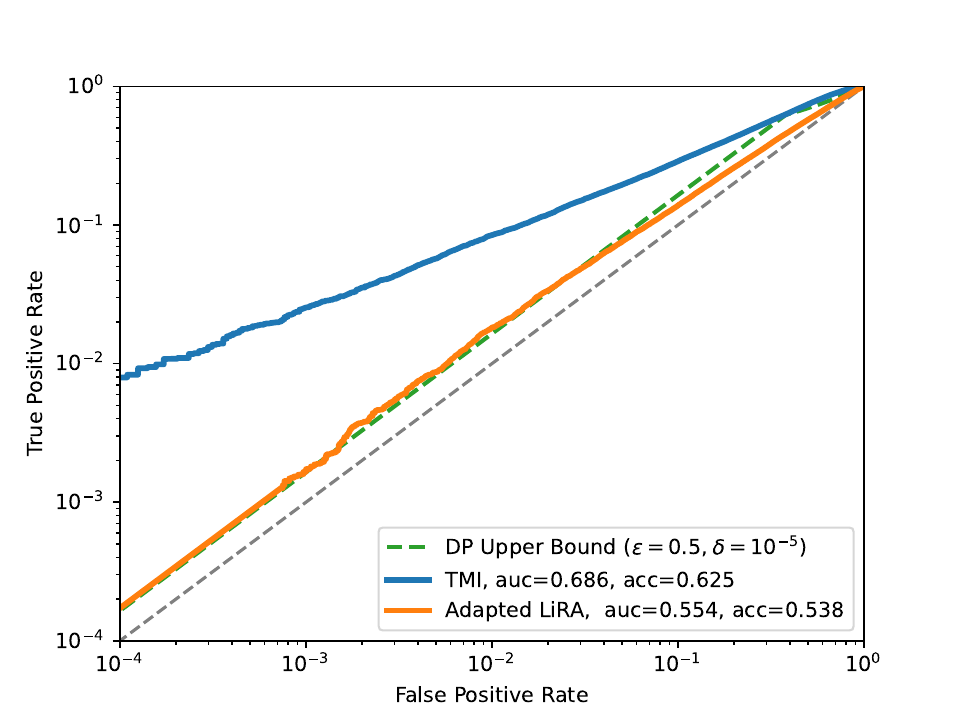}
      \caption{$\varepsilon = 0.5, \delta=10^{-5}$}
    \end{subfigure}
    \begin{subfigure}[t]{.49\linewidth}
      \includegraphics[width=\linewidth]{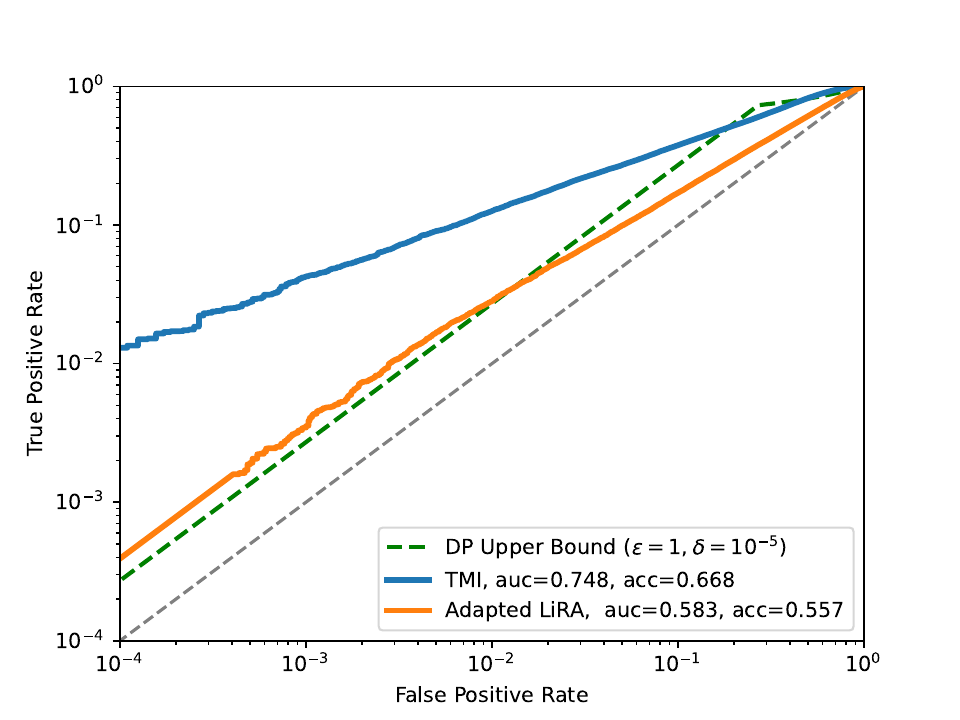}
      \subcaption{$\varepsilon = 1, \delta=10^{-5}$}
    \end{subfigure}
    \captionsetup{justification=centering}
    \caption{\revision{\attackname\ Performance on Samples Present in Both $D_{PT}$ \textit{and} $D_{FT}$ when Finetuning Models with DP-SGD}}
    \label{fig:overlap}
\end{figure*}

\revision{
While it may seem expected that finetuning on a disjoint dataset with DP-SGD provides no privacy guarantee for individuals in the pretraining set, the authors of \cite{2022-tramer-position} pose the following question: \textit{What privacy guarantee should an individual expect if their data was present in both pretraining and finetuning?} This scenario is not unlikely, as large models are trained on terabytes of data scraped from the Web \cite{2020-gpt3, JFT300}. Because manually inspecting these datasets is infeasible, it is likely that an individual's datapoint which was included in private finetuning is also present in \textit{non-private} pretraining. Thus, their data does not enjoy the $(\varepsilon, \delta)-DP$ guarantee promised by finetuning, as the corresponding pretraining gradients are unbounded in magnitude and exact in direction. Misusing DP-SGD in this manner can leave these individuals at risk of privacy attacks.

To support our claim that finetuning with DP-SGD is blatantly non-private when $D_{PT} \cap D_{FT} \neq \emptyset$, we run experiments on the CIFAR-100 dataset. Similar to our prior experiments, we finetune the final layer of ResNet-34 shadow models on the Coarse CIFAR-100 dataset using DP-SGD~\cite{abadi-2016-dpsgd} with target privacy parameters $\varepsilon \in \{0.5, 1\}$ and $\delta=10^{-5}$ and clipping norm equal to 5. The only difference in this experiment is the fact that the finetuning set contains some ($\sim$1000) individuals who were also present in the pretraining task. We run our \attackname\ attack on these individuals and report the results in Figure~\ref{fig:overlap} and Table~\ref{tab:overlap}. When we finetune with DP-SGD and the challenge points are included in the finetuning set, we see true positive rates that are comparable to our experiments on models finetuned using feature extraction (Table~\ref{tab:tpr-feature-extraction}). At a fixed FPR of 0.1\% and target privacy guarantees of $\varepsilon=0.5$ and $\varepsilon=1$, \attackname\ achieves a TPR 15.6$\times$ higher than the upper bound (end-to-end) training with DP-SGD should provide.  
}
% imply anything about the overall privacy of the model. 

\noindent \textbf{\ref{rq:privacy} Answer}: Finetuning a pretrained model using DP-SGD provides a privacy guarantee \textit{only} for the downstream dataset. Therefore, DP-SGD has little to no impact on privacy risk of the pretraining dataset, and these downstream models leak the membership status of individuals in the pretraining dataset.  

\revision{In settings where the pretraining and finetuning data overlap, the guarantee that differential privacy typically provides does not hold. This happens because any given individual's influence in the pretraining process is unbounded and deterministic. Thus, pretraining can induce leakage of individuals in the finetuning set, even when DP-SGD is used to finetune the model.}

\begin{table}[h]
\centering
\caption{TPR at Fixed FPR of \attackname\ when Target Models are Finetuned with DP-SGD (Figure~\ref{fig:dpsgd})}
\label{tab:tpr-dp}
\resizebox{.65\columnwidth}{!}{%
\begin{tabular}{@{}lcc@{}}
\toprule
\multicolumn{1}{c}{\textbf{Task}}     & \textbf{TPR @ 0.1\% FPR} & \textbf{TPR @ 1\% FPR} \\ \midrule
Coarse CIFAR-100 ($\varepsilon=\infty$) & 5.7\%  & 16.1\% \\
Coarse CIFAR-100 ($\varepsilon=1.0, \: \delta=10^{-5}$)        & 3.2\%  & 10.6\% \\
Coarse CIFAR-100 ($\varepsilon=0.5, \: \delta=10^{-5}$)       & 3.3\%  & 10.7\% \\
CIFAR-10 ($\varepsilon=\infty$) & 2.0\%  & 8.0\%  \\
CIFAR-10 ($\varepsilon=1.0, \: \delta=10^{-5}$)                & 1.6\%  & 6.6\%  \\
CIFAR-10 ($\varepsilon=0.5, \: \delta=10^{-5}$)                & 2.1\%  & 6.6\%   \\ \bottomrule \\ 
\end{tabular}%
}

\end{table}

\begin{table}[h]
\centering
\caption{\revision{TPR at Fixed FPR of \attackname\ Performance on Samples Present in Both $D_{PT}$ \textit{and} $D_{FT}$ when Finetuning Models with DP-SGD (Figure~\ref{fig:overlap})}}
\label{tab:overlap}
\resizebox{.65\columnwidth}{!}{%
\begin{tabular}{@{}lcc@{}}
\toprule
\multicolumn{1}{c}{\textbf{Task}}     & \textbf{TPR @ 0.1\% FPR} & \textbf{TPR @ 1\% FPR} \\ \midrule
\attackname\ ($\varepsilon=0.5, \: \delta=10^{-5}$)       & 2.5\%  & 8.5\% \\ 
\attackname\ ($\varepsilon=1.0, \: \delta=10^{-5}$)        & 4.2\%  & 12.6\% \\ 
Theoretical Upper Bound ($\varepsilon=0.5, \: \delta=10^{-5}$)       & 0.16\%  & 1.7\% \\
Theoretical Upper Bound ($\varepsilon=1.0, \: \delta=10^{-5}$)        & 0.27\%  & 2.7\% \\ \bottomrule \\ 
\end{tabular}%
}

\end{table}
\section{Discussion and Conclusion}

% Main Takeaways:
% \begin{enumerate}
%     \item Transferred models leak private information about their pretraining sets via black box queries
%     \item Whether you use feature extraction or finetune the model's pretrained parameters, privacy leakage is still present
%     \item  Even when the tasks are very different, there is still some amount of privacy leakage with respect to the pretraining set
%     \item Using differential privacy on a downstream task, even with aggressive privacy parameters, does not inhibit privacy risk with respect to the pretraining dataset
%     \item Pretrained foundation language models may be susceptible to privacy leakage. 
% \end{enumerate}

We study the critical issue of privacy leakage in the transfer learning setting by proposing a novel threat model and introducing \attackname, a metaclassifier-based membership-inference attack. In particular, we explore how finetuned models can leak the membership status of individuals in the pretraining dataset without an adversary having direct access to the pretrained model. Instead, we rely on queries to the finetuned model to extract private information about the pretraining dataset. 

Through our evaluation of \attackname, we demonstrate privacy leakage in a variety of transfer learning settings. We demonstrate the effectiveness of our attack against a variety of models in both the vision and natural language domains, highlighting the susceptibility of finetuned models to leaking private information about their pretraining datasets. In the vision domain, we show that \attackname\ is effective at inferring membership when the target model is finetuned using various strategies, including differentially private finetuning with stringent privacy parameters. We also demonstrate the success of our attack on publicly hosted foundation models by adapting \attackname\ to use a global metaclassifier.

\paragraph{Other Privacy Attacks on Finetuned Models}
% Property Inference, Attribute Inference, Extraction
We introduce the first threat model that uses query access to a finetuned model to mount a privacy attack on pretraining data. It remains an open question as to whether other privacy attacks, such as property inference, attribute inference, and training data extraction attacks can also see success in this transfer learning setting. Given that MI attacks are used as practical tools to measure or audit the privacy of machine learning models~\cite{song-2018-auditingmi, tensorflow-2020-auditing, ye-2022-enhanced_mi}, future work should consider efficiency and simplicity when designing new privacy attacks in the transfer learning setting.

\paragraph{Considerations for Private Machine Learning}
Our evaluation shows that the pretraining dataset of machine learning models finetuned with differential privacy are still susceptible to privacy leakage. This supports the argument made in \cite{2022-tramer-position} that "privacy-preserving" models derived from large, pretrained models don't necessarily provide the privacy guarantees that consumers of services backed by these finetuned models would expect. Prior works that utilize public data to improve the utility of differentially private machine learning models have made strides towards making differential privacy practical for several deep learning tasks \cite{papernot-2020-making, yu-2022-dpllm, li-2022-llmstrongdp, bu-2023-dpbiasft, he-2022-groupwisedp, ganesh-2023-pretraining,golatkar-2022-mixeddp}, but they do not address privacy risks external to model training itself.  

Using \attackname\ as a measurement of privacy leakage in this setting, we reinforce the fact that maintaining privacy depends on taking a holistic approach to the way that training data is handled. As stated in~\cite{2022-tramer-position}, privacy is not binary (i.e. not all data is either strictly "private" or "public") and privacy in machine learning is not only dependent on the model's training procedure. To grapple with privacy risk in this increasingly popular transfer learning setting, researchers and practitioners should explore new ways to sanitize sensitive information from training datasets of machine learning models, create ways to collect potentially sensitive Web data with informed consent from individuals, and work towards end-to-end privacy-preserving machine learning with high utility and privacy guarantees.

% Fine-tuning with DP doesn't make a model private with respect to the pretraining set. DP ML should consider more end-to-end solutions to help with the privacy-utility tradeoff (e.g. mixed public private training).

% \myparagraph{Future Work}
% % Making more efficient attacks, getting closer to ROC upper bound of lira

\clearpage
\printbibliography

\appendix
\revision{
\section{Membership Inference Under Distribution Shift} \label{adx:theoretical}

We analyze the success of membership inference attacks on Gaussian mean estimation under distribution shift. While this setting is simple compared to the finetuned deep learning models, it helps us understand how repurposing one estimator for a new problem can leak information about the original dataset.

\subsection{Introduction}

In this setting, the challenger has access to two datasets, $X$ and $Y$, where $\vert X \vert \gg \vert Y \vert$. The challenger uses these datasets to publish a statistic that is a combination of the empirical means of each dataset. This can be thought of as leveraging the the larger dataset, $X$, to estimate a statistic that comes from a similar distribution. The adversary's goal is the following: Given a challenge point $c$, determine the membership status of $c$ with respect to the dataset $X$. 

Let $X = \{x_1, \dots, x_n\}$ and $Y = \{ y_1, \dots, y_m \}$ be datasets where each $x_i \sim \mathcal{N}(\mu, \mathbb{I}_d)$ and $y_i \sim \mathcal{N}(\mu + \nu, \mathbb{I}_d)$, and let $\hat{\mu} = \alpha \Bar{x} + (1 - \alpha) \Bar{y}$ be the statistic released by the challenger. 

Then, 
\begin{align*}
    \ex{}{\hat{\mu}} = \mu + (1 - \alpha) \nu
\end{align*}
and
\begin{align*}
    \cov{}{\hat{\mu}} &= \alpha^2 \var{}{\Bar{x}} + (1-\alpha)^2 \cov{}{\Bar{y}}  \\ 
    &= \frac{\alpha^2}{n} \cdot \mathbb{I}_d + \frac{(1-\alpha)^2}{m} \cdot \mathbb{I}_d \\
    &= \Big( \frac{\alpha^2}{n} + \frac{(1-\alpha)^2}{m} \Big) \cdot \mathbb{I}_d \\
    &= \Tilde{\alpha} \cdot \mathbb{I}_d
\end{align*}

% \ju{Should analyze $\ex{}{\| \hat\mu - \mu \|^2}$.}

In this mean estimation setting, $\norm{\nu}{2}$ can be thought of as the inversely proportional to the similarity between the pretraining and finetuning tasks. If $\mu$ is similar to the mean of the new data, $Y$, $\norm{\nu}{2}$ is small. The term, $\alpha$, is analagous to the fraction of pretraining epochs (i.e. number of pretraining epochs divided by the total number of pretraining and finetuning epochs). For example if there are 80 pretraining epochs and 20 finetuning epochs, the corresponding $\alpha$ value would be $0.8$. Note that as $\alpha \to 0$, the information from the empirical mean of $X$ is completely overshadowed by the empirical mean of $Y$. Prior work on membership-inference attacks on machine learning models has suggested that gradient updates (a special case of mean estimation) that do not contain an individual make membership-inference success decrease with respect to that individual \cite{jagielski-2023-forgetting}. This is consistent with the results we present in this section for the simplified setting of membership-inference attacks on Gaussian mean estimation.

\captionsetup[subfigure]{width=0.9\textwidth}
\begin{figure*}[!h]
\centering
    \begin{subfigure}[t]{.49\textwidth}
      \includegraphics[width=0.9\textwidth]{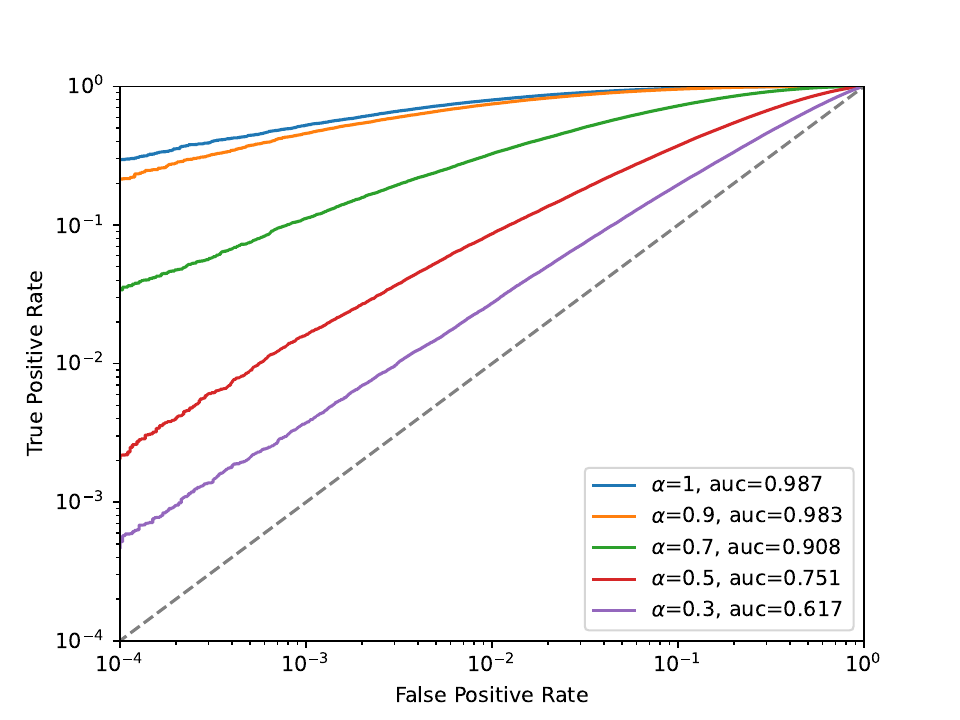}
      \caption{\revision{ROC of the membership inference attack on mean estimation for various values of $\alpha$ ($d=10,000, \: n=1000, \: m=100)$}}
    \end{subfigure}
    \begin{subfigure}[t]{.49\textwidth}
      \includegraphics[width=0.9\textwidth]{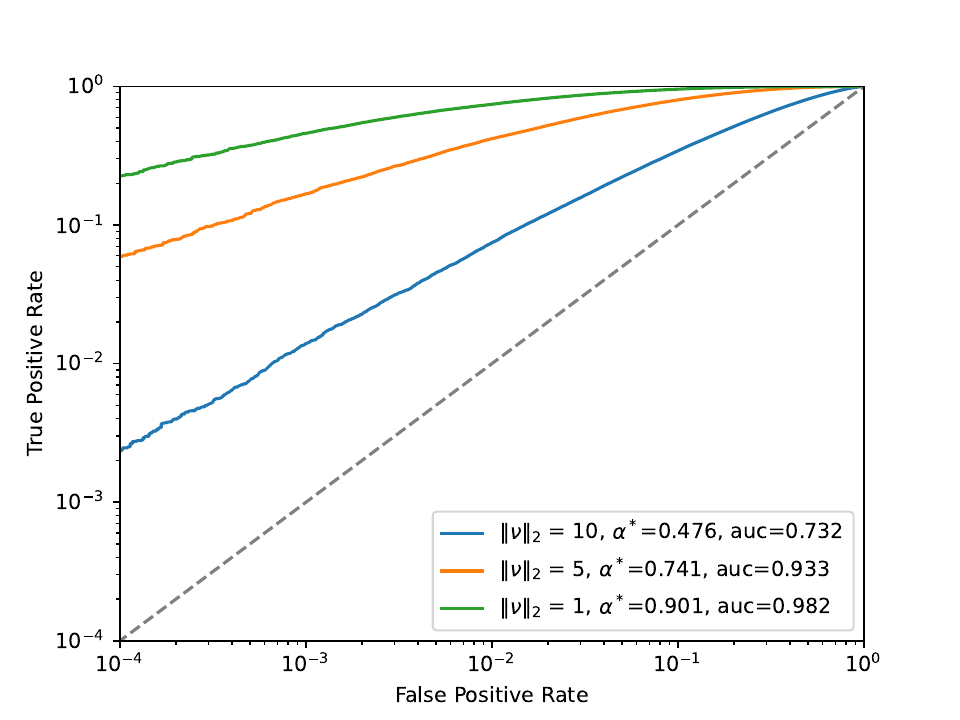}
      \subcaption{\revision{ROC of the membership inference attack on mean estimation for varying amounts of distribution shift, $\norm{\nu}{2}$. In each of these simulations, $\alpha$ is set optimally according to Lemma~\ref{lem:opt-alpha}}}
    \end{subfigure}
    \captionsetup{justification=centering}
    \caption{\revision{Performance of the Membership Inference Attack on Mean Estimation}}
    \label{fig:theoretical_roc}
\end{figure*}

\subsection{Threat Model and Attack Algorithm}

Similar to prior work on membership inference-attacks on mean estimation, we assume that the adversary has query access to the aggregate statistic, $\hat{\mu}$, along with the true mean of this statistic, $\ex{}{\hat{\mu}}$. The membership-inference security game between the challenger and the adversary is defined as the following:

\begin{enumerate}
    \item Pick $b \sim \cU(\{0,1\})$
    \item If $b = 0$, sample the challenge point, $c \sim \cN(\mu, \mathbb{I}_d)$, else sample $c$ uniformly from $X$
    \item Compute $z = \inn{\hat{\mu} - \ex{}{\hat{\mu}}}{c - \ex{}{c}} = \inn{\hat{\mu} - (\mu + (1 - \alpha) \nu)}{c - \mu}$
    \item If $z > \tau$, output $1$. Else, output $0$
\end{enumerate}

\subsection{Results} \label{adx:results}

\begin{lem}
    If $c$ is OUT ($b=0$), then 
    \[
        \ex{}{z} = 0 \quad \textrm{and} \quad \var{}{z} = d \Tilde{\alpha},
    \] 
    and if $c$ is IN ($b=1$), then 
    \[
        \ex{}{z} = \frac{\alpha d}{n} \quad \textrm{and} \quad \var{}{z} = d \Tilde{\alpha} + \frac{2d \alpha^2}{n^2}
    \]
\end{lem}

\begin{proof}
    We will begin by analyzing the \textit{OUT} case, where $c \sim \mathcal{N}(\mu, \mathbb{I}_d)$

    \begin{align*}
        \ex{}{z} &= \ex{}{\inn{\hat{\mu} - (\mu + (1 - \alpha) \nu)}{c - \mu}} \\ 
        &= \inn{\ex{}{\hat{\mu} - (\mu + (1 - \alpha) \nu)}}{\ex{}{c - \mu}} \\
        &= \inn{\Vec{0}}{\Vec{0}} \\
        &= 0
    \end{align*}

    Next, we compute the variance in the \textit{OUT} case:

    \begin{align*}
        \var{}{z} &= \var{}{\inn{\hat{\mu} - (\mu + (1 - \alpha) \nu)}{c - \mu}} \\ 
        &= \sum_{i=1}^{d}{\var{}{\inn{\hat{\mu} - (\mu + (1 - \alpha) \nu)}{c - \mu}}}\\
        &= \sum_{i=1}^{d}{\ex{}{(\hat{\mu} - (\mu + (1 - \alpha) \nu))^{2}_{i} \cdot (c - \mu)^{2}_{i}}} \\
        &= \sum_{i=1}^{d}{\ex{}{(\hat{\mu} - (\mu + (1 - \alpha) \nu))^{2}_{i}} \cdot \ex{}{ (c - \mu)^{2}_{i}}} \\
        &= \sum_{i=1}^{d}{\Tilde{\alpha} \cdot 1} \\ 
        &= d \Tilde{\alpha}
    \end{align*}

    Now, we will analyze the IN case, where $c$ is sampled uniformly at random from the dataset, $X$. In this case, the published statistic, $\hat{\mu}$ and the challenge point $c$ are \textit{not} independent. For succinctness, let $t = (\mu + (1 - \alpha) \nu)$. 

    \begin{align*}
        \ex{}{z} &= \ex{}{\inn{\hat{\mu} - t}{c - \mu}} \\ 
        &= \sum_{i=1}^{d}{\ex{}{(\hat{\mu} - t)_i \cdot (c - \mu)_i}} \\ 
        &= \sum_{i=1}^{d}{\ex{}{\hat{\mu}_i \cdot c_i} - \mu_i \cdot \ex{}{\hat{\mu}_i} -t_i \cdot \ex{}{c_i} + \mu_i t_i} \\
        &= \sum_{i=1}^{d}{\ex{}{\hat{\mu}_i \cdot c_i} - \mu_i \cdot \ex{}{\hat{\mu}_i}} \\
        &= \sum_{i=1}^{d}{\ex{}{\hat{\mu}_i \cdot c_i} - \mu_i t_i} \\
    \end{align*}

    \noindent Since $c = x_{i} \in X$ for some $i$, without loss of generality suppose $c_i = x_{i, 1}$ for all $i$. Then, $\ex{}{\hat{\mu}_i \cdot c_i}$ becomes

    \begin{align*}
        \ex{}{\hat{\mu}_i \cdot c_i} &= \ex{}{c_i \cdot \frac{\alpha}{n} \sum_{j=1}^{n}{ x_{i, j}} + c_i \cdot (1-\alpha) \Bar{y}_i} \\ 
        &= \ex{}{\frac{\alpha}{n} \cdot x_{i, 1} \cdot c_i + \frac{\alpha}{n} \sum_{j=2}^{n}{ c_{i} \cdot  x_{i, j}} + c_i \cdot (1-\alpha) \Bar{y}_i} \\ 
        &= \ex{}{\frac{\alpha}{n} \cdot x_{i, 1} \cdot c_i} + \ex{}{\frac{\alpha}{n} \sum_{j=2}^{n}{c_{i} \cdot x_{i, j}} + c_i \cdot (1-\alpha) \Bar{y}_i} \\ 
        &= \ex{}{\frac{\alpha}{n} \cdot x_{i, 1}^{2}} + \ex{}{\frac{\alpha}{n} \sum_{j=2}^{n}{ c_{i} \cdot x_{i, j}} + c_i \cdot (1-\alpha) \Bar{y}_i} \\ 
        &= \frac{\alpha}{n}\ex{}{x_{i, 1}^{2}} + \frac{\alpha }{n} \sum_{j=2}^{n}{\ex{}{c_{i}} \cdot \ex{}{x_{i, j}}} + \ex{}{c_i} \cdot (1-\alpha) \ex{}{\Bar{y}_i} \\ 
        &= \frac{\alpha}{n}(\mu_{i}^{2} + 1) + \frac{\alpha }{n} \sum_{j=2}^{n}{\mu_{i}^{2}} + \mu \cdot (1-\alpha) (\mu + \nu)_i \\ 
        &= \frac{\alpha}{n}(\mu_{i}^{2} + 1) + \frac{\alpha (n - 1)}{n} \mu_{i}^{2} + \mu \cdot (1-\alpha) (\mu + \nu)_i \\ 
    \end{align*}

    \noindent Plugging the above terms into the original expression for the expectation of $\ex{}{z}$ and simplifying yields

    \begin{align*}
        \sum_{i=1}^{d}{\ex{}{\hat{\mu}_i \cdot c_i} - \mu_i t_i} &= \sum_{i=1}^{d}{\frac{\alpha}{n}} \\
        &= \frac{\alpha d}{n}
    \end{align*}

    \noindent Lastly, we compute the variance in the \textit{IN} case:

    \begin{align*}
        \var{}{z} &= \var{}{\inn{\hat{\mu} - t}{c - \mu}} \\ 
        &= \sum_{i=1}^{d}{\var{}{(\hat{\mu} - t)_i \cdot (c - \mu)_i}} \\ 
        &= \sum_{i=1}^{d}{\var{}{\hat{\mu}_i \cdot c_i - \mu_i \cdot \hat{\mu}_i - t_i \cdot c_i }} \\
        % &= \sum_{i=1}^{d}{\var{}{\hat{\mu}_i \cdot c - \mu \cdot \hat{\mu} - t \cdot c}} \\
    \end{align*}

    \noindent Since $c = x_{i} \in X$ for some $i$, without loss of generality suppose $c_i = x_{i, 1}$ for all $i$. For succinctness, we drop the summation and indices, $i$, since all of the dimensions are i.i.d. Expanding $\hat{\mu}$, we get

    \[
    \var{}{x_1 \cdot \big( \frac{\alpha}{n}  \sum_{j=1}^{n}{x_j} + (1-\alpha) \Bar{y} \big)  - \mu \cdot \big( \frac{\alpha}{n}  \sum_{j=1}^{n}{x_j} + (1-\alpha) \Bar{y} \big)  - t \cdot x_1}
    \]

    \noindent We will also use the shorthand $\beta = \frac{\alpha}{n}  \sum_{j=2}^{n}{x_j} + (1-\alpha) \Bar{y}$. Note that $\beta$ is normally distributed with mean $\mu + (1-\alpha) \nu = t$ and variance $\frac{\alpha^2 (n-1)}{n^2} + \frac{(1-\alpha)^2}{m}$. Pulling $x_1$ out of the summations yields
    
    \begin{align*}
        &= \var{}{\frac{\alpha}{n} x_{1}^2 + x_{1} \beta  - \frac{\alpha \mu}{n}  \cdot x_{1} - \mu \beta  - t \cdot x_1}  \\
        &= \var{}{\frac{\alpha}{n} x_{1}^2 + x_{1} (\beta  - \frac{\alpha \mu}{n} -t) - \mu \beta} \\
        &= \ex{}{(\frac{\alpha}{n} x_{1}^2 + x_{1} (\beta  - \frac{\alpha \mu}{n} -t) - \mu \beta)^{2}} \\
        &- \ex{}{(\frac{\alpha}{n} x_{1}^2 + x_{1} (\beta  - \frac{\alpha \mu}{n} -t) - \mu \beta)}^{2}
    \end{align*}

    After algebraic manipulation and computing the individual expectations as in the OUT case, we arrive at

    \begin{align*}
        &= \sum_{i=1}^{d}{\Tilde{\alpha} + \frac{2\alpha^{2}}{n^{2}}}  \\
        &= d \Tilde{\alpha} + \frac{2d \alpha^{2}}{n^{2}}
    \end{align*}
    
\end{proof}

% \ju{We can calculate the square of the ratio of the difference in expectation to the standard deviation as approximately
% \[
%     \frac{d}{n} \cdot \frac{\alpha^2 m }{\left(
%     \alpha^2 m + (1-\alpha)^2 n \right)}
% \]}

\begin{lem}
    The mean squared error of $\hat{\mu}$ (as an estimator of the mean of the finetuning data, $\mu + \nu$) is the following:
    \[
        \ex{}{\|\hat{\mu} - (\mu + \nu)\|^{2}} = \Tilde{\alpha}d + \alpha^2 \norm{\nu}{2}^{2}
    \]
    The choice of $\alpha$ that minimizes the mean-squared-error is 
    \[
        \alpha^{*} = \frac{d}{m(\norm{\nu}{2}^2 + \frac{d}{n}) +  d }
    \]
\end{lem} \label{lem:opt-alpha}

\begin{proof}
    Consider the mean squared error of $\hat{\mu}$ as an estimator of the mean of the finetuning data , $\mu + \nu$.
    \[
        \ex{}{\|\hat{\mu} - (\mu + \nu)\|^{2}}
    \]
    
    Note that $Z = \hat{\mu} - (\mu + \nu) \sim \mathcal{N}(-\alpha \nu, \Tilde{\alpha} \mathbb{I}_d)$ and for any multivariate Gaussian random variable, $X \sim \mathcal{N}(\mu, \Sigma)$, we have 
    \[
        \ex{}{\|X\|^{2}} = Tr(\Sigma) + \| \mu \|^{2} 
    \]
    Thus, the mean squared error is 
    \[
        \ex{}{\|\hat{\mu} - (\mu + \nu)\|^{2}} = \Tilde{\alpha}d + \alpha^2 \norm{\nu}{2}^{2}
    \]
    Suppose the challenger who is releasing $\hat{\mu}$ wants to choose $\alpha$ (i.e. the pretraining-to-finetuning split) such that the error on the finetuning data, $Y$, is minimized. Computing the derivative of the mean squared error with respect to $\alpha$ yields

    \[
    MSE'(\alpha) = 2d \Big( \frac{\alpha}{n} - \frac{1-\alpha}{m} \Big) + 2\alpha \| \nu \|
    \]
    
    Setting $MSE'(\alpha) = 0$ and solving for $\alpha$, we find that the optimal parameter, $\alpha^{*}$, is
    \[
        \alpha^{*} = \frac{d}{m(\norm{\nu}{2}^2 + \frac{d}{n}) +  d }
    \]
    
\end{proof}

% Assumes that z is Gaussian
% Should not be a lemma

\begin{lem}
    Assume the test statistic, $z$, is normally distributed. The AUC of our membership-inference attack can be written as the probability the test statistic, $z$, for an \textit{IN} sample exceeds $z$ for an \textit{OUT} sample:
    \[
        AUC  = \frac{1}{2} \Bigg( 1 + \texttt{erf} \bigg(\frac{\alpha d}{2\sqrt{d ( \Tilde{\alpha}n^{2} + \alpha^{2}}) } \bigg) \Bigg) 
    \]
\end{lem}

\begin{proof}
The AUC of a classifier can be thought of as the probability that the prediction value on a random positive example exceeds the prediction value on a random negative example.
    \[
        AUC = \pr{}{z_{IN} > z_{OUT}}
    \]
where $z_{IN} \sim \mathcal{N}(\frac{\alpha d}{n}, d \Tilde{\alpha} + \frac{2d \alpha^2}{n^2})$ and $z_{OUT} \sim \mathcal{N}(0, d \Tilde{\alpha})$. Subtracting the the two random variables, we get
    \begin{align*}
        AUC &= \pr{}{z_{IN} > z_{OUT}} \\
        &= \pr{}{z_{OUT} - z_{IN} < 0} \\ 
        &= \pr{}{\mathcal{N} \Big( -\frac{\alpha d}{n}, \: 2d\Tilde{\alpha} + \frac{2 d \alpha^{2}}{n^{2}} \Big) < 0} \\ 
        &= \frac{1}{2} \Bigg(1 + \texttt{erf} \Bigg( \frac{\frac{-\alpha d}{n}}{\sqrt{2}\cdot \sqrt{(2d\Tilde{\alpha} + \frac{2d\alpha^{2}}{n^{2}})}} \Bigg) \Bigg) \\ 
        &= \frac{1}{2} \Bigg( 1 + \texttt{erf} \bigg(\frac{\alpha d}{2\sqrt{d ( \Tilde{\alpha}n^{2} + \alpha^{2}}) } \bigg) \Bigg) 
    \end{align*}
\end{proof}
}

\newpage

\section{Ablations}

In this section, we evaluate variations of \attackname. We limit the adversary's access to the target model's prediction outputs, consider different choices of metaclassifier architecture, and study how the number of challenge point queries affects the effectiveness of our attack.  

\begin{figure}[h!]
    \centering
    \includegraphics[width=0.5\textwidth]{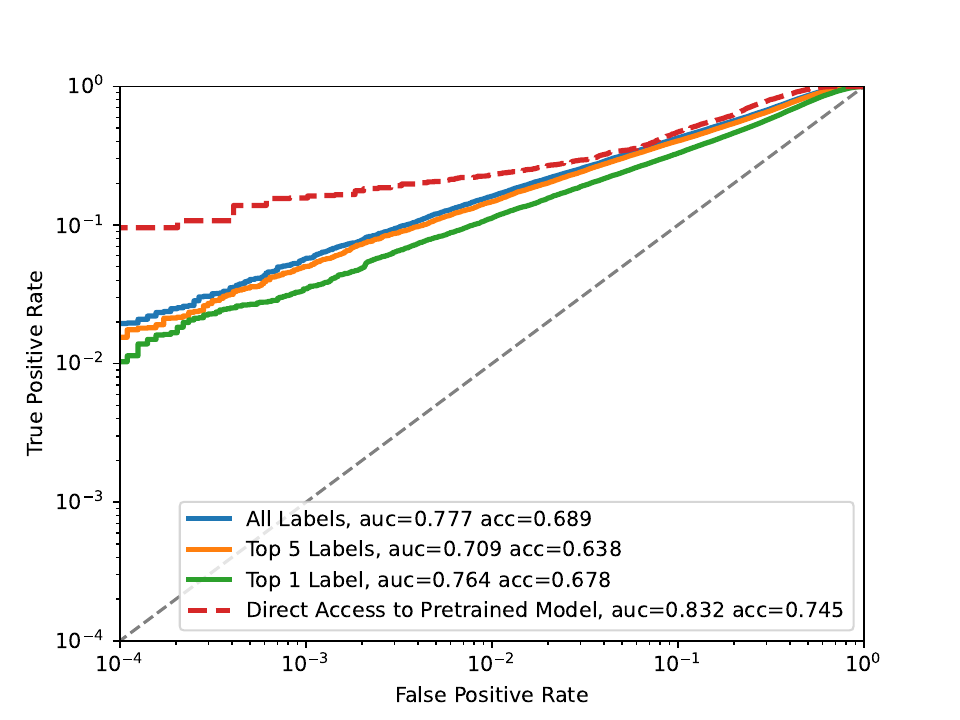}
    \caption{\attackname\ Performance with Access to Prediction Confidence on the Top-K Labels}
    \label{fig:topk}
\end{figure}

\subsection{Access to Top-k Predictions} \hfill

In many realistic settings, an adversary who has query access to a computer vision model may only receive predictions for top-k most probable labels. Because our attack relies the information from a combination of labels, we evaluate \attackname\ with access to the top 1, top 5, and all labels in the downstream task. For this experiment, we use the same pretrained and finetuned models as in our experiments with Coarse CIFAR-100. This time, when we query the shadow models and target model, we mask the prediction confidences on all but the top-k labels. Because the prediction confidences always sum up to 1, we take the remaining probability mass and divide it amongst the remaining labels to construct the vectors for the metaclassifier (e.g. if the top 5 predictions make up 0.90 of the total probability mass, we divide 0.10 across the remaining 15 labels).

In Figure~\ref{fig:topk}, we show the performance of \attackname\ when the adversary has access to the top 1, 5, and 20 labels in our Coarse CIFAR-100 task. Interestingly, \attackname\ with access to a single label has higher attack success than our adaptation of LiRA (Figures~\ref{fig:feature_extraction-c100} and \ref{fig:feature_extraction-tiny}) even though both adversaries are given the same amount of information. This may be due to the fact that we create some additional information about the other classes by constructing a prediction vector using the labels that the adversary has access to, which is only possible if the adversary knows all of the possible class labels a priori.

\subsection{Different Metaclassifier Architectures} \hfill

Throughout our evaluation, we primarily use a neural network as our metaclassifier to perform our membership-inference attack. In this ablation, we study how the choice of metaclassifier affects the success of our attack. We use the following architectures: neural network multilayer perceptron, support vector machine, logistic regression, and k-nearest-neighbor ($k=\sqrt{D_{\text{meta}}}$). When using the k-nearest-neighbor metaclassifier, we receive hard-label (binary) predictions for membership status. This stands in contrast ot the continuous scores that we receive from the three other metaclassifier architectures. To obtain a membership score in the interval $[0,1]$, we average the labels from the k-nearest-neighbor models across the prediction vectors obtained from several different augmented queries to the target model. Although we receive a continuous score from each of the other metaclassifiers, we also average the scores across all of the prediction vectors from augmented queries.

\begin{figure}[h!]
    \centering
    \includegraphics[width=0.5\textwidth]{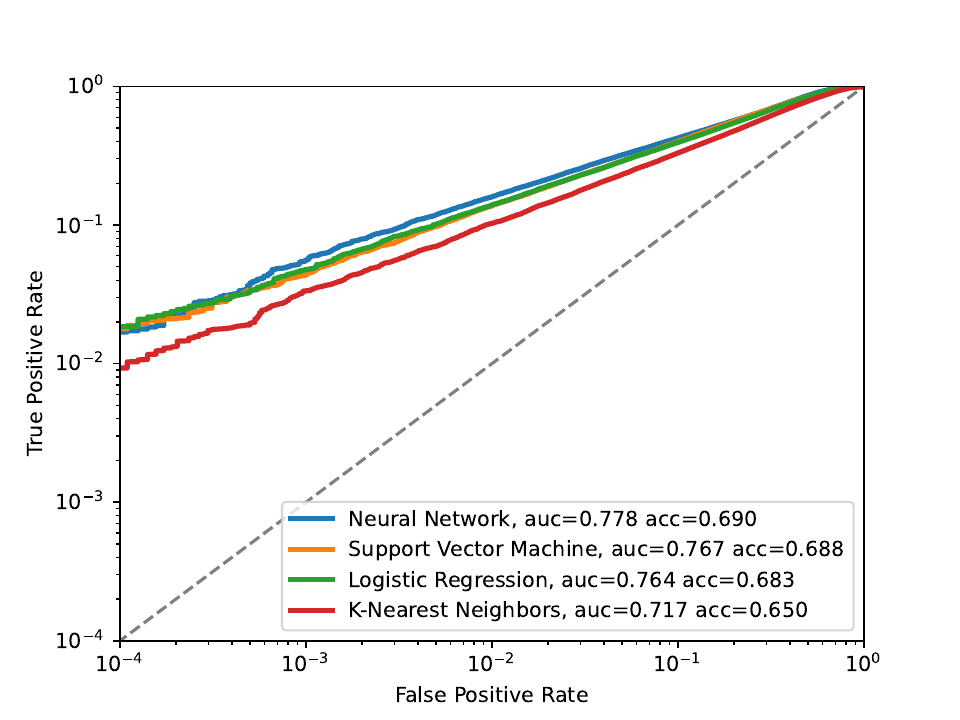}
    \caption{\attackname\ Performance with Different Metaclassifier Architectures}
    \label{fig:architectures}
\end{figure}

We observe that the overall impact on the effectiveness of our attack is minimal, which indicates that \attackname\ is relatively robust to the choice of metaclassifier architecture. Figure~\ref{fig:architectures} shows that the AUC and accuracy slightly decrease when using metaclassifiers other than a neural network. This suggests that an adversary could potentially use faster metaclassifiers than neural networks, such as logistic regression and k-nearest neighbors, without significantly compromising the effectiveness of the attack.

\subsection{Number of Augmented Queries} \hfill

Our attack relies on several augmented queries of a single challenge point on a handful of local shadow models to construct a sufficiently sized metaclassifier dataset. We also query the target model on these augmentations and average the metaclassifier predictions. In this experiment, we explore how the number of augmentations of a challenge point affects the success of \attackname.

\begin{figure}[h!]
    \centering
    \includegraphics[width=0.5\textwidth]{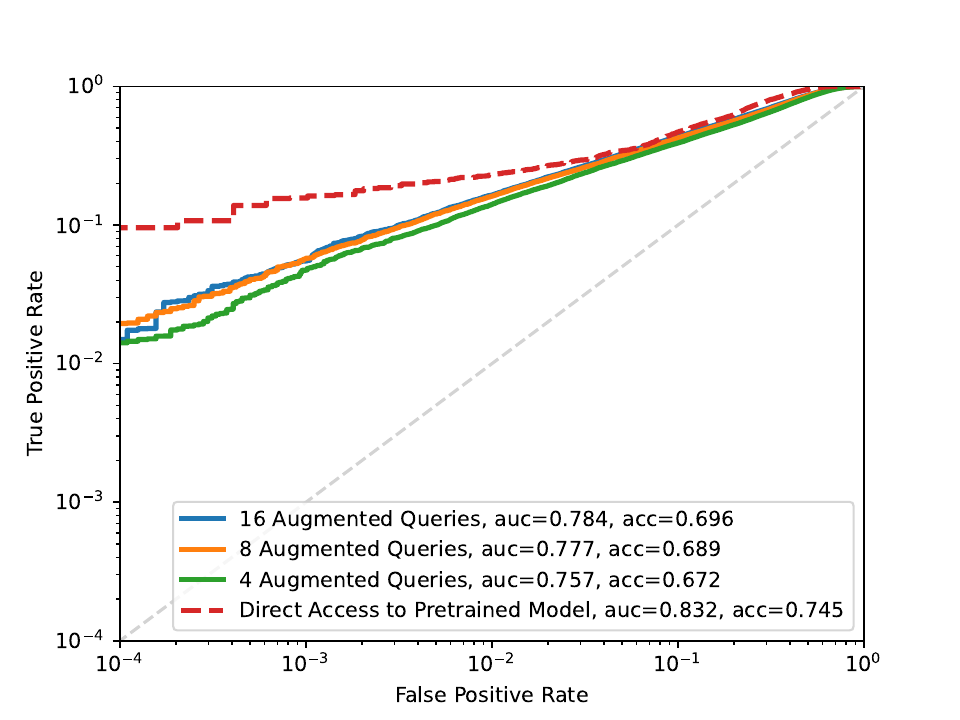}
    \caption{\attackname\ Performance with Different Numbers of Augmented Queries}
    \label{fig:augmentations}
\end{figure}

Figure~\ref{fig:augmentations} shows that using more augmentations increases the FPR, AUC, and balanced accuracy of our attack. Although \attackname\ is more effective with a higher number of augmented queries, training metaclassifiers becomes increasingly computationally expensive as $D_{\text{meta}}$ becomes large. For example, \attackname\ runs 6$\times$ slower on our hardware when using 16 augmentations of the challenge point instead of 8. In all of our prior experiments, we use 8 augmentations to strike a balance between attack effectiveness and efficiency.

\newpage

\section{Datasets and Models}

\revision{In this section, we discuss each of the datasets and the training procedures used in our evaluation of \attackname.}

\subsection{Datasets} \label{sec:datasets}

\begin{itemize}
    
    \item \textit{CIFAR-100: } The CIFAR-100~\cite{CIFAR} dataset is a subset of the Tiny Images dataset \cite{TinyImages}, provided by the Canadian Institute for Advanced Research. It is comprised of 60,000 32x32 color images from 100 classes, where each class contains 600 images (500 for training and 100 for testing). CIFAR-100 is used as one of our pretraining tasks because it is a challenging dataset with a wide variety of classes, which allows our models to learn very general features and patterns that can be applied to several downstream tasks.

    \item \revision{\textit{Tiny ImageNet:}  Tiny ImageNet \cite{tiny-imagenet} is an image classification dataset designed to be a smaller scale alternative to the popular ImageNet \cite{imagenet} dataset. This dataset contains 110,000 64x64 color images from 200 classes, where each class contains 550 images (500 for training and 50 for testing). We use Tiny ImagenNet to pretrain the larger Wide ResNet architecture because it provides a similarly general task to CIFAR-100 at a larger scale. Additionally, the full-sized version of ImageNet is a widely used dataset for pretraining large image models, thus reinforcing the need to evaluate our attack on a dataset like Tiny ImageNet.}
    
    \item \textit{Coarse CIFAR-100: } The classes in CIFAR-100 can be divided into 20 superclasses. Each image in the dataset has a "fine" label to indicate its class and a "coarse" label to indicate its superclass. We construct this coarse dataset using the superclass labels and use it as our downstream task with the highest similarity to the pretraining task. In our experimentation, we ensure that this downstream task does not contain any of the pretraining samples from the standard CIFAR-100 dataset.
    
    \item \revision{\textit{Caltech 101:} Caltech 101 \cite{caltech-101} is an image classification dataset comprised of about 9000 color images from 101 classes, where each class contains 40 to 800 images. Because the images in this dataset vary in size and tend to be high resolution, we downscale them to be 64x64 to reduce computational complexity. We use the Caltech 101 dataset to finetune our pretrained Tiny ImageNet models as it provides a difficult task with many categories that can be solved by leveraging the generic features learned during pretraining.}

    \item \textit{Oxford-IIIT Pet: } The Oxford-IIIT Pet Dataset~\cite{pets} is made up of about 7400 color images of cats and dogs. This dataset contains 37 classes with roughly 200 images per class. In our evaluation, this downstream task is the least similar to CIFAR-100 because it focuses on a specific category of images that are mutually exclusive to the pretraining set classes. For this task, we do not use feature extraction because the finetuned model have low utility. Rather, we use the pretrained model as an initialization and update all of its weights. 
    
    % \revision{\item \textit{Tiny ImageNet}}
    \item \textit{Caltech 101: } In a similar fashion to CIFAR-100, the CIFAR-10~\cite{CIFAR} is comprised of 60,000 32x32 color images selected from the Tiny Images dataset. This dataset includes 10 classes, each containing 6000 points (5000 for training and 1000 for testing) which are mutually exclusive to those seen in CIFAR-100. In our evaluation, this downstream task is the second most similar to CIFAR-100 because they are both derived from the same distribution of web-scraped images, but they are disjoint in their classes. Although the classes do not overlap, the features learned from pretraining on CIFAR-100 may be useful in performing this task.
    
    \item \textit{WikiText-103: } WikiText-103~\cite{wikitext103} is a large-scale language dataset that is widely used for benchmarking language models. It contains over 100 million tokens and is derived from several Wikipedia articles and contains a vast amount of textual data. The language models we consider in this paper have been pretrained on the train partition of WikiText-103 and are hosted on Hugging Face.
    
    % The "103" in the dataset name refers to the approximate number of million tokens present in the dataset. The dataset is preprocessed in a way that retains the original text structure and includes a wide range of topics and writing styles.
    
    \item \textit{DBpedia:} The DBpedia ontology (or topic) classification dataset~\cite{dbpedia} is composed of 630,000 samples with 14 non-overlapping classes from DBpedia, which is a project aiming to extract structured content from the information on Wikipedia. For each of the 14 topics, there are 40,000 training samples and 5000 testing samples. In our experiments with language models, we update a subset of the model's weights on random subsets of this dataset.

    \revision{\item \textit{Yahoo Answers: } The Yahoo Answers topic classification dataset~\cite{2015-yahoo} is composed of 1.4 million training samples and 60,000 testing samples with 10 classes. The training and testing sets are divided evenly amongst the topics, such that there are 140,000 training samples and 6000 testing samples per class. Each sample contains both the title and content of a question asked on Yahoo Answers. In our experiments with language models, we update a subset of the model's weights on random subsets of this dataset, where the content of each question is appended to the question title.}
\end{itemize}

\revision{
\subsection{Shadow Model Training} \label{adx:sm_training}

\revision{Our shadow model training procedure for vision models is split into two phases: pretraining and finetuning. In the first phase, we train 129 randomly initialized ResNet models on random subsets of Tiny ImageNet and CIFAR-100, each containing half of the dataset (50k and 25k points, respectively). The remaining samples are held out for evaluation. We train each of the ResNet-34 models for 100 epochs (to 75-80\% top-5 accuracy) using SGD with weight decay ($\lambda=10^{-5}$) and cosine annealing \cite{loshilov-2017-annealing} as our learning rate scheduler. Using the same hyperparameters, we train the Wide ResNet-101 architecture for 200 epochs to 60\% top-5 accuracy. When training and querying any of our shadow models, we use standard data augmentations, such as random crops and horizontal flips.

In the second phase, we finetune our shadow models on randomly sampled subsets of our downstream task datasets. Before we finetune each shadow model, we swap the classification layer out with a randomly initialized one that has the proper dimension for the downstream task. We then freeze a subset of the model's pretrained weights. When we use feature extraction to finetune our pretrained models, we freeze all weights except for those in the final classification layer. The weights that aren't frozen are trained using the same process as pretraining, but for 20\% of the epochs. 

When pretraining our shadow models, we designate a randomly selected set of points to be the challenge points for our \attackname\ attack. Because each shadow model is trained on half of the dataset, all of the points (including the challenge points) will be IN and OUT for approximately half of the shadow models. In our experiments, we select one shadow model to be the target model and run our attack using the remaining 128 shadow models. Each time we run our attack, we select the a different shadow model to be the target model, yielding a total of 128 trials.}
}

\end{document}